\documentclass[journal]{IEEEtran}

\usepackage{graphicx} 
\usepackage{booktabs, multirow}
\usepackage{amsfonts}
\usepackage{amsmath}
\usepackage[hang,flushmargin]{footmisc}
\usepackage[hyphens]{url}
\usepackage[hidelinks]{hyperref}
\usepackage{microtype}
\usepackage{algorithm}
\usepackage{algorithmic}
\usepackage{xcolor}
\renewcommand{\arraystretch}{0.85}
\begin{document}

\title{Toward Interpretable Multimodal Fusion: Heat Conduction Modeling for Hyperspectral and LiDAR Joint Classification}

\author{Kan Wei,
        Jiahui Cui,
        Jing Yao,~\IEEEmembership{Senior Member,~IEEE},
        Xinyu Zhao,
        Lei Wang,~\IEEEmembership{Member,~IEEE},
        and Pedram Ghamisi,~\IEEEmembership{Senior Member,~IEEE}%


\thanks{Kan Wei, Jiahui Cui, Xinyu Zhao and Lei Wang are with the Aerospace Information Research Institute, Chinese Academy of Sciences, Beijing 100190, China, the Key Laboratory of Target Cognition and Application Technology, Beijing 100190, and the School of Electronic, Electrical and Communication Engineering, University of Chinese Academy of Sciences, Beijing 100190, China (e-mails: weikan24@mails.ucas.ac.cn; cuijiahui24@mails.ucas.ac.cn; zhaoxinyu@aircas.ac.cn; wanglei002931@aircas.ac.cn).}

\thanks{Jing Yao is with the State Key Laboratory of Remote Sensing and Digital Earth, Aerospace Information Research Institute, Chinese Academy of Sciences, Beijing 100094, China (e-mail: yaojing@aircas.ac.cn).}

\thanks{P. Ghamisi is with the Helmholtz-Zentrum Dresden-Rossendorf, 09599 Freiberg, Germany, and also with the Faculty of Electrical and Computer Engineering, University of Iceland, 101 Reykjavik, Iceland (e-mail: p.ghamisi@gmail.com).}}


\markboth{IEEE Transactions on Circuits and Systems for Video Technology}{Wei \MakeLowercase{\textit{et al.}}}

\IEEEpubid{%
\parbox{\textwidth}{\centering\scriptsize
Copyright \copyright~20xx IEEE. Personal use of this material is permitted.
However, permission to use this material for any other purposes must be obtained
from the IEEE by sending an email to pubs-permissions@ieee.org.}}

\maketitle

\begin{abstract}

The fusion of hyperspectral (HS) and Light Detection and Ranging (LiDAR) data plays a crucial role in enhancing land-cover classification by jointly exploiting spectral, spatial, and structural cues. However, existing multimodal fusion methods still struggle to model long-range dependencies and complex anisotropic interactions while maintaining computational efficiency. This paper introduces M2Heat, a physics-inspired framework that investigates multimodal fusion through the lens of heat conduction. At its core, a physics-driven visual heat conduction module (vHeat) and enhanced Frequency Value Embeddings (FVEs) simulate anisotropic information flow, enabling the capture of global dependencies with sub-quadratic complexity and physical interpretability. This mechanism, combined with a hybrid spatial-frequency fusion strategy named Cross-Frequency Fusion (CFF) module, produces highly discriminative and robust feature representations. M2Heat achieves competitive overall performance on three benchmarks, i.e., Trento, Houston2013, and Augsburg, while providing an interpretable heat-conduction-guided perspective for multimodal feature fusion. These results indicate the potential of heat-conduction-guided neural operators for efficient and interpretable RS multimodal fusion. The source code is publicly available at \url{https: /github.com/Weikan0425/M2Heat_HSI_LiDAR}.

\end{abstract}


\IEEEpeerreviewmaketitle

\section{Introduction}

\IEEEPARstart{I}{n} recent years, Earth observation (EO) technologies have played an increasingly vital role in monitoring land surface dynamics over large areas and in near real-time \cite{chen2025cangling,wei2026beyond}. Among the various remote sensing modalities, hyperspectral images (HSI) have emerged as a particularly powerful tool due to their exceptionally high spectral resolution \cite{hsi-review}. Compared to conventional optical sensors such as RGB or multispectral images (MSI), HSI captures hundreds of contiguous spectral bands, enabling detailed characterization of materials based on their unique spectral signatures. This capability has been widely demonstrated to be effective in a broad range of applications, including land cover mapping \cite{weiMGFNetMLPdominatedGated2024}, \cite{xie2025fsg}, vegetation analysis \cite{hsi-veg-1}, \cite{hsi-veg-2}, ecological monitoring \cite{sun2025mask}, \cite{hsi-em-2}, and mineral exploration \cite{hsi-soil}.

\begin{figure}
    \centering
    \includegraphics[width=0.8\linewidth]{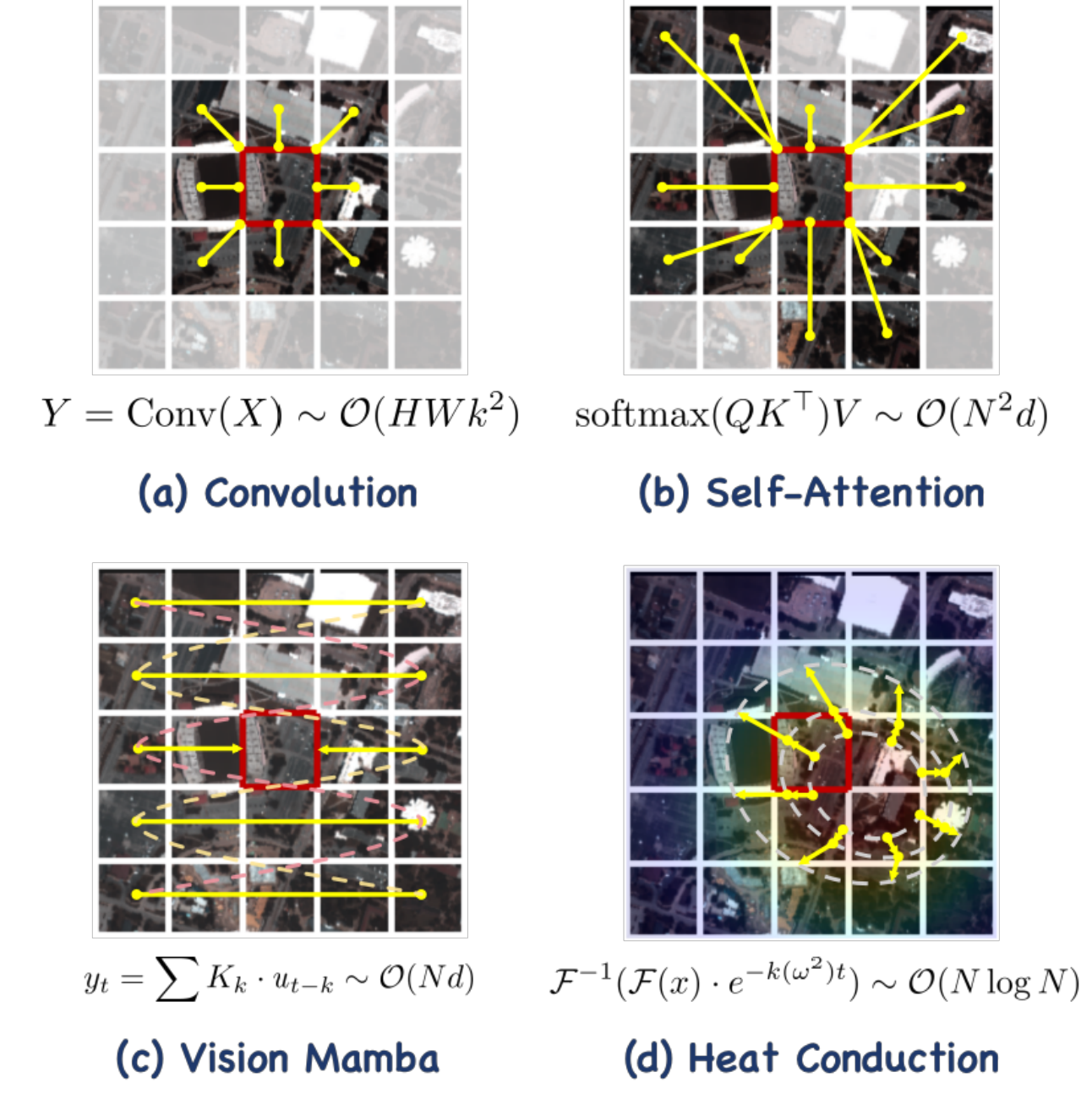}
    \caption{Comparison of mainstream modeling mechanisms. (a) Convolution. (b) Self-Attention. (c) Vision Mamba. (d) Heat Conduction.}
    \label{fig:fig1}
\end{figure}

\IEEEpubidadjcol

Recent advances in deep learning (DL) have substantially improved hyperspectral image (HSI) classification by enabling more effective modeling of spatial and spectral dependencies~\cite{songInteractiveSpectralspatialTransformer2024},~\cite{liuSpatialawareConformalPrediction2025}. Convolutional neural networks (CNNs) first demonstrated strong capabilities in extracting local spatial–spectral patterns~\cite{zhaoJointClassificationHyperspectral2023},~\cite{wangS3F2NetSpatialSpectralStructuralFeature2025}, while transformer-based architectures further enhanced global context modeling through self-attention mechanism~\cite{zhangMultimodalInformativeViT2024},~\cite{maRoSENetRotationSimilarity2025}. More recently, state-space models such as Mamba have introduced efficient long-range dependency modeling~\cite{mambahsi},~\cite{mamba-2}, further extending the representational power of DL-based HSI methods. 

Nevertheless, single-modality approaches still face challenges in complex real-world environments with diverse land covers, shadow effects, or structural occlusions~\cite{xuMultisourceRemoteSensing2018}. To address these limitations, multimodal remote sensing (RS) has emerged as a promising paradigm by integrating complementary information from heterogeneous sensors such as HSI and LiDAR, which respectively provide fine spectral and rich structural cues~\cite{hongMoreDiverseMeans2021},~\cite{stf}.

Driven by the progress of multimodal DL, a variety of fusion-based architectures have been proposed to jointly exploit spectral and geometric representations~\cite{songMCFNetMultiscaleCrossDomain2025}. While early works relied on simple concatenation or shared-branch designs, modern methods explore adaptive attention-based~\cite{wangCITNetConvolutionInteraction2024} and hybrid fusion frameworks~\cite{yuHI2D2FNet2023},~\cite{dongJointContextualRepresentation2023}. Despite these advances, achieving robust and well-aligned multimodal fusion remains challenging due to modality heterogeneity and scale disparities. This motivates the development of a unified, dynamically adaptive fusion framework for multimodal RS classification.

To address the aforementioned limitations, this work explores two complementary directions: developing an interpretable global modeling mechanism and designing an effective multimodal fusion framework. As shown in Fig.~\ref{fig:fig1}, convolutional operations are efficient for local representation but struggle to capture long-range dependencies, whereas self-attention provides global context modeling at the cost of quadratic complexity. Recently, state-space models such as Mamba~\cite{gu2023mamba,zhu2024unetmamba} have achieved linear-complexity global modeling, yet their interpretability remains limited for structured remote sensing data. 

Motivated by this gap, we introduce a physics-inspired framework, termed M2Heat, which performs global sub-quadratic and interpretable modeling through a heat conduction operator (HCO) that simulates information diffusion across spatial–spectral dimensions. Building upon the vHeat paradigm~\cite{wangBuildingVisionModels}, M2Heat extends heat-based modeling to the multimodal hyperspectral–LiDAR domain by predicting learnable thermal diffusivity rates via Frequency Value Embeddings (FVEs). Furthermore, we design a Cross-Frequency Fusion (CFF) module that aligns and fuses heterogeneous spectral and geometric representations in the frequency domain. Together, these designs establish a unified framework that jointly achieves efficient global modeling and semantically consistent multimodal fusion for remote sensing classification.

The main contributions can be summarized as follows:

\begin{itemize}
    \item[\textbf{1)}] A physics-driven multimodal framework, termed M2Heat, is established to achieve efficient HSI–LiDAR joint classification. At its core lies a cross-modal Fourier fusion mechanism that integrates modality-specific and shared representations within a unified spectral–spatial domain. By embedding diffusion dynamics into deep representations, this framework models long-range spatial–spectral dependencies with sub-quadratic complexity, thereby achieving a principled balance between interpretability and computational efficiency.

    \item[\textbf{2)}] Central to the design is the vHeat module equipped with FVEs, which reformulates anisotropic heat propagation in the frequency domain, where FVEs act as physics-guided priors regulating the diffusion behavior of spectral–spatial features. Such formulation enhances discriminative feature learning and improves model generalization across heterogeneous modalities.

    \item[\textbf{3)}] We introduce a hybrid cross-domain fusion mechanism, combining alignment-aware interaction in the spatial domain with spectral energy redistribution in the frequency domain. Through this dual-phase integration, complementary modality-specific cues and shared semantic information are adaptively aligned and fused, yielding a structurally consistent multimodal representation.

    \item[\textbf{4)}] Extensive evaluations conducted on three public benchmarks, i.e., Augsburg, Houston2013, and Trento, demonstrate the competitive performance and practical effectiveness of M2Heat. The ablation studies validate the effectiveness of FVEs and the hybrid fusion strategy. This study explores the potential of interpretable heat-diffusion dynamics for HSI--LiDAR classification, bridging physical modeling and deep multimodal representation learning.
\end{itemize}

\section{Related work}
\subsection{Feature Extraction and Representation Learning: From Local to Global}

\textbf{CNN-based Models:} 
Convolutional Neural Networks (CNNs) have long served as the foundation for HSI classification due to their ability to hierarchically capture local spatial–spectral structures. By stacking convolutional layers, CNNs progressively extract discriminative features that effectively characterize land-cover patterns.  
To extend CNNs to multimodal scenarios, various studies have explored joint exploitation of HSI spectral and LiDAR structural information. Representative examples include the 3D CNN-based H+L framework~\cite{royHyperspectralLiDARData2022}, the multiscale MSNetSC~\cite{xueMultiscaleDeepLearning2022}, and the triplet-based TSDN~\cite{liTripletSemisupervisedDeep2022}. In addition, single-stream CNNs~\cite{yangSingleStreamCNNLearnable2022} have been proposed to jointly learn modality-shared features without explicit separation. More recently, DHNet~\cite{jin2025dhnet} constructed spatial-specific and spectral-specific heterogeneous branches with feature calibration to enhance complementary spectral--spatial representation learning for HSI classification.
  
Despite their effectiveness in local representation, CNN-based methods are inherently constrained by fixed receptive fields, limiting their capacity to model long-range dependencies critical for multimodal understanding.

\vspace{2pt}
\textbf{Transformer-based Models:} 
Transformers have recently gained prominence in HSI analysis by effectively modeling global contextual dependencies~\cite{hongSpectralFormerRethinkingHyperspectral2022}. Their extension to multimodal fusion has led to architectures such as DHViT~\cite{xueDeepHierarchicalVision2022} and MFT~\cite{royMultimodalFusionTransformer2023a}, which employ cross-attention for inter-modal correlation learning. Hybrid networks such as GLT-Net~\cite{dingGlobalLocalTransformer2022} further integrate CNNs and Transformers to balance local and global modeling, while efficient variants like ExViT~\cite{yaoExtendedVisionTransformer2023} and SWFormer~\cite{liSWFormerStochasticWindows2024} reduce computational cost through lightweight attention designs.  
Nevertheless, Transformer-based models remain computationally expensive and often lack modality awareness, motivating research into more efficient and interpretable multimodal modeling approaches.

\vspace{2pt}
\textbf{Mamba-based Models:} 
Recently, State Space Models (SSMs) have emerged as efficient alternatives to self-attention for long-sequence modeling. The Mamba architecture, in particular, achieves linear complexity while retaining strong global modeling ability~\cite{wangMambaHSIMultidirectionalState2025}.  
Its adaptation to multimodal RS tasks has shown promising results, such as HLMamba~\cite{liaoJointClassificationHyperspectral2024} and S$^2$CrossMamba~\cite{zhangS2CrossMambaSpatialSpectral2024}, which enhance semantic representation and cross-modal feature fusion. Subsequent works like SSFN~\cite{luoVMambaBasedSpatialSpectral2025} and M2FMNet~\cite{panMultimodalFusionMamba2025} further unify spatial and spectral dependencies under Mamba-based frameworks.  
Despite their efficiency, current Mamba-based methods rely on sequential tokenization of 2D images, which weakens spatial interpretability and motivates the exploration of more physically grounded modeling paradigms.

\vspace{2pt}
\textbf{Physics-inspired Models:} 
Incorporating physical priors such as thermodynamics and diffusion processes into deep learning models has proven effective for improving interpretability and structural consistency. These priors introduce inductive constraints that align model behavior with real-world physical principles, which is particularly valuable for RS applications involving energy transport and material interactions.

Recent studies have explored physics-guided paradigms to enhance multimodal feature alignment and semantic consistency. For example, Yu et al.~\cite{yuHI2D2FNet2023} integrated physically constrained embeddings into HSI–LiDAR fusion, while Li et al.~\cite{liFedDiffDiffusionModel2024} proposed a diffusion-based framework inspired by nonequilibrium thermodynamics. Wang et al.~\cite{wangBuildingVisionModels} further introduced vHeat, a heat-diffusion-based model achieving global context modeling with sub-quadratic complexity, and Hu et al.~\cite{huRSvHeatHeatConduction2025} extended it to RS-vHeat for spatial diffusion in optical and SAR imagery. From the broader perspective of interpretable multimodal representation, UAAFusion~\cite{bai2025uaafusion} formulated multimodal image fusion as an attribution-analysis-guided deep unfolding process to improve task-aware interpretability. FAMAFuse~\cite{kamara2026famafuse} introduced a functional--anatomical multiscale attention mechanism to adaptively balance modality-specific details and global contextual information in multimodal image fusion.

\begin{figure*}
    \centering
    \includegraphics[width=0.85\linewidth]{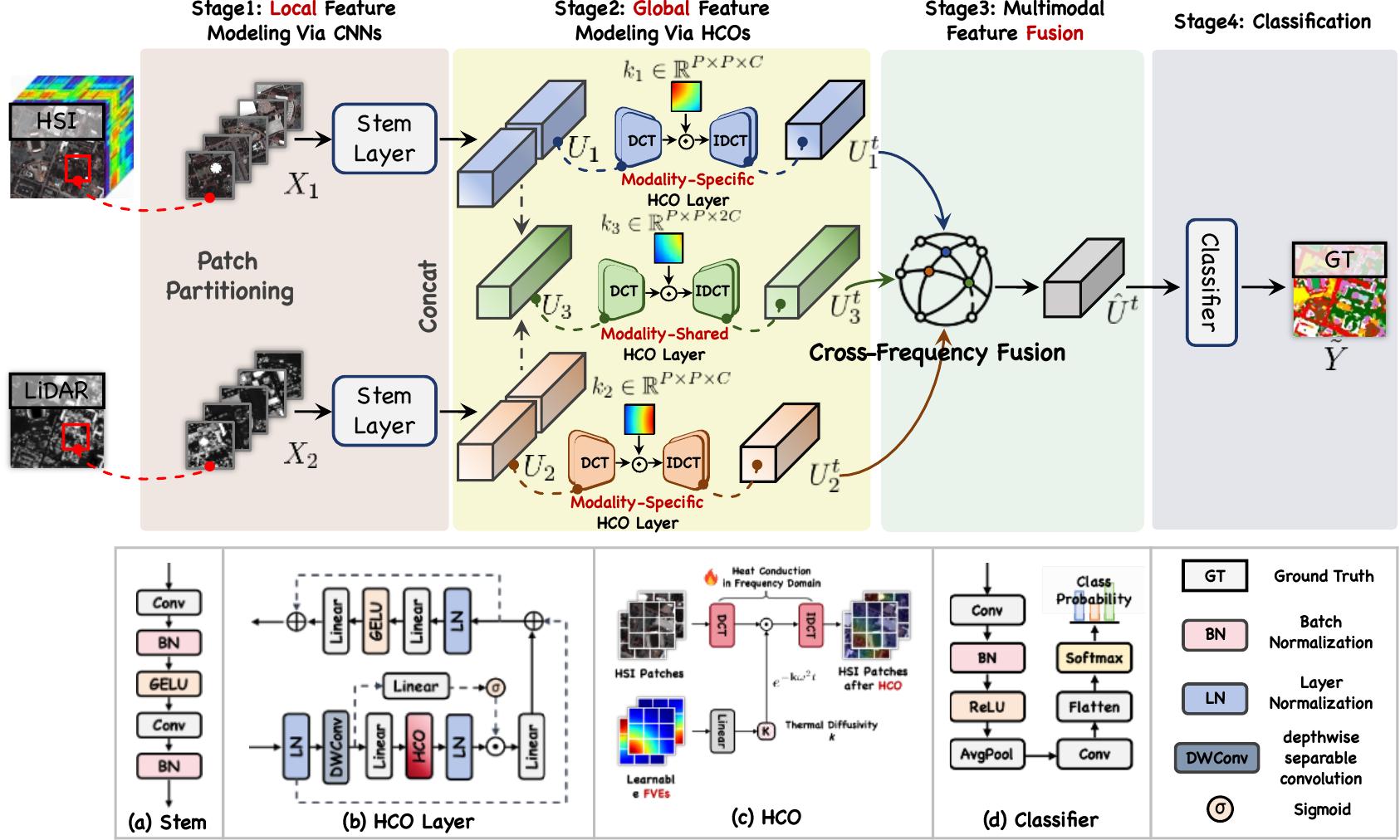}
    \caption{Overall framework of the proposed M2Heat, which consists of Stem Layer, Modality-specific and Modality-shared HCO Layer, CFF module, and classifier.}
    \label{fig:fig2}
\end{figure*}
  
In diffusion-process-inspired modeling, diffusion models typically formulate diffusion as an iterative generative or refinement process, whereas vHeat and RS-vHeat discretize heat conduction as a deterministic feature-propagation operator for visual and RS representation learning. RS-vHeat further demonstrates the potential of heat-conduction modeling in optical and SAR-oriented remote sensing scenarios. For HSI--LiDAR joint classification, heterogeneous spectral signatures and LiDAR-derived geometric structures require coupled spectral--spatial--structural propagation and cross-modal alignment. Accordingly, M2Heat refines heat-conduction-guided modeling to the HSI--LiDAR fusion scenario by introducing modality-specific and shared HCO branches with cross-frequency fusion, aiming to achieve interpretable and efficient multimodal representation learning.

\subsection{Multimodal Alignment and Feature Fusion: From Shallow Interaction to Hybrid Integration}

The integration of heterogeneous modalities, such as HSI and LiDAR, has proven highly effective for enhancing scene understanding in RS. However, challenges persist due to significant disparities in spatial resolution, data distribution, and semantic structure across modalities. To address these, early approaches focused on feature alignment and shallow interaction. For instance, Hong et al.~\cite{hongMultimodalGANsCrossmodal2021} proposed SM-GANs, introducing an early cross-fusion strategy to capture complementary cues in urban scenes. Hang et al.~\cite{hangCrossModalityContrastiveLearning2022} designed a cross-modality contrastive learning method that enables unsupervised feature alignment without labeled data. CCR-Net~\cite{wuConvolutionalNeuralNetworks2022} further unified alignment and fusion via a cross-channel reconstruction module, enhancing inter-modality correlation.

Recent efforts have shifted toward hierarchical and hybrid fusion strategies to capture deeper semantic dependencies. Flex-MCFNet~\cite{wangMultistageInformationComplementary2024} sequentially applies information complement and global fusion modules for stage-wise integration. PID-HLfusion~\cite{yuPIDHLfusionPluggableProgressive2024} leverages parameter sharing and geometric priors to achieve progressive modality fusion. CCEnd-Net~\cite{daiCCEndNetCrossModalCascaded2025} adopts a cascaded encoder–decoder design for early-to-late fusion, while DCMNet~\cite{linDynamicCrossModalFeature2025} employs multi-level routing spaces to dynamically align and integrate spatial–spectral features. MIViT~\cite{zhang2024mivit} introduced an information aggregation--distribution mechanism to enhance complementary feature interaction while reducing redundant multimodal information for HSI--LiDAR classification.
S3F2Net~\cite{wang2025s3f2net} developed a spatial--spectral--structural feature fusion framework to jointly exploit HSI spectral--spatial characteristics and LiDAR-derived structural cues.
In parallel, Qin et al.~\cite{qinCollaborativeClassificationHyperspectral2025} explored fractional Fourier transforms and optimal matching flows for effective cross-modal interaction.

Together, these advances underscore a clear trend toward unified, adaptive, and context-aware frameworks capable of bridging modality gaps through deeper and more flexible interactions. Nonetheless, further exploration is required to fully exploit the complementary information of heterogeneous data.

\section{Methodology}

\subsection{Overall Framework of M2Heat}
As illustrated in Fig.~\ref{fig:fig2}, the proposed M2Heat framework is a physically inspired multimodal architecture designed to jointly model HSI and LiDAR data. The framework aims to simulate the thermal diffusion process in a visual–spectral context, thereby achieving efficient cross-modal information propagation and structurally consistent feature alignment. Specifically, given a pair of co-registered HSI and LiDAR patches, M2Heat first extracts shallow modality representations via Stem Layers, followed by modality-specific Heat Conduction Operator (HCO) Layers that perform physics-inspired spatial–spectral diffusion. Subsequently, a shared HCO Layer is used for cross-modal alignment, ensuring that heterogeneous representations are projected into a unified latent space. A Cross-Frequency Fusion (CFF) module then facilitates multi-frequency information interaction, and a lightweight classifier produces the final pixel-wise semantic predictions. The overall framework couples physical diffusion dynamics with deep feature learning, offering an interpretable multimodal feature-learning framework for HSI--LiDAR classification.

\subsection{Multimodal Feature Extraction of M2Heat}
\subsubsection{Physical and Visual Motivation}
The core design of M2Heat is grounded in the classical heat conduction equation, which describes the spatio-temporal evolution of temperature in a continuous medium. Let $u(x, y, t)$ denote the temperature distribution at position $(x, y)$ and diffusion time $t$ over a 2D domain $\mathbf{D} \subset \mathbb{R}^2$. The fundamental governing equation can be expressed as:
\begin{equation}
    \frac{\partial u}{\partial t} = k \left( \frac{\partial^2 u}{\partial x^2} + \frac{\partial^2 u}{\partial y^2} \right),
\end{equation}
where $k>0$ represents the thermal diffusivity coefficient controlling the rate of energy transfer. The analytical solution under initial condition $u(x,y,0)=f(x,y)$ can be derived via the 2D Fourier Transform:
\begin{equation}
    \tilde{u}(\omega_x, \omega_y, t) = \tilde{f}(\omega_x, \omega_y)e^{-k(\omega_x^2+\omega_y^2)t},
\end{equation}
which reveals that each frequency component $\tilde{f}(\omega_x, \omega_y)$ is exponentially attenuated according to its spatial frequency magnitude. This behavior implies that low-frequency components representing global smooth structures are preserved, while high-frequency details gradually decay, forming a physically grounded low-pass diffusion effect. The heat conduction process thus naturally provides a mathematical model for information propagation, structural smoothing, and multi-scale feature evolution.

Motivated by this principle, we reinterpret visual feature maps as temperature fields, where each channel represents an energy distribution evolving through a learnable diffusion process. The evolution of a feature field $U(x,y,c,t)$ can be approximated as:
\begin{equation}
    U_t = \text{IDCT}_{2D}\left(\text{DCT}_{2D}(U_0)\cdot e^{-k(\omega_x^2+\omega_y^2)t}\right),
\end{equation}
where $\text{DCT}_{2D}$ and $\text{IDCT}_{2D}$ denote the 2D Discrete Cosine Transform and its inverse, respectively. Compared to the traditional Fourier transform, DCT naturally satisfies Neumann (reflective) boundary conditions, avoiding boundary artifacts and making it well-suited for finite image patches.  
This formulation enables M2Heat to implement heat diffusion efficiently in the spectral domain, achieving a balance between local smoothness and global consistency. The computational complexity of this diffusion process is approximately $\mathcal{O}(N^{1.5})$, ensuring scalability for large-scale RS data.

\begin{figure*}
    \centering
    \includegraphics[width=0.75\linewidth]{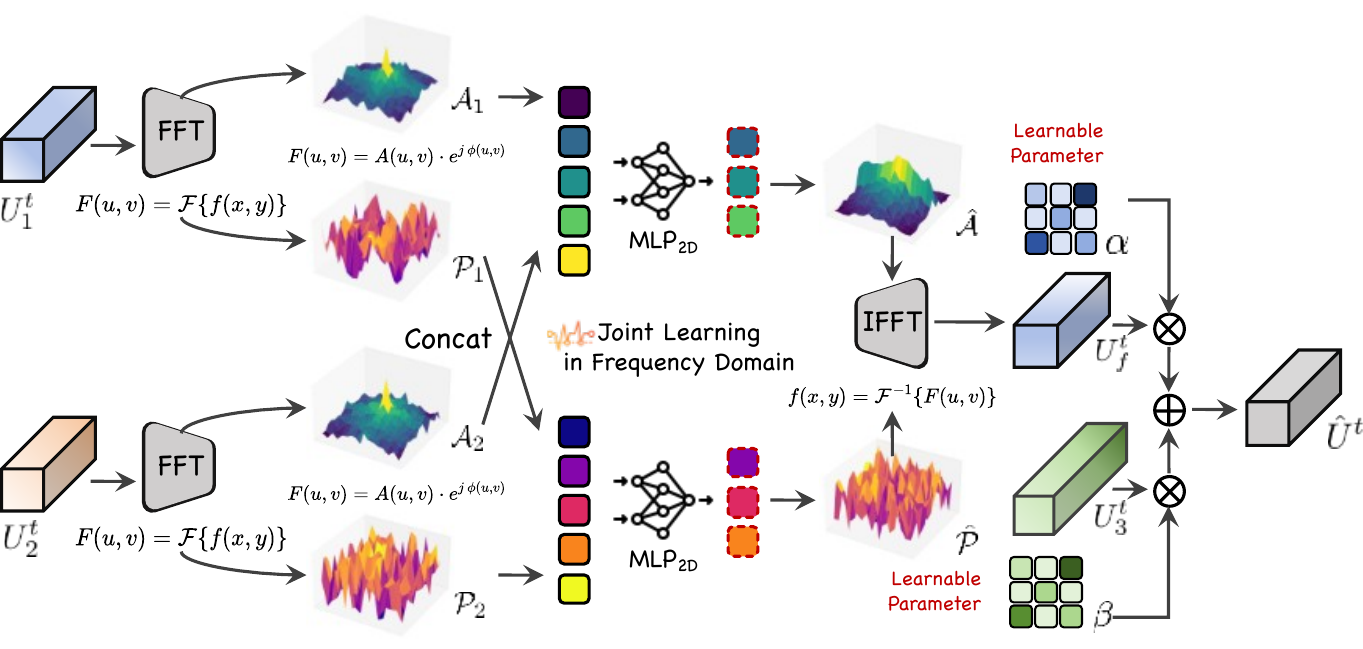}
    \caption{Our proposed CFF module, which integrates fusion in the frequency domain and dynamic fusion with jointly learned features, enabling effective multimodal feature interaction and fusion.}
    \label{fig:fig4}
\end{figure*}

\subsubsection{Heat Conduction Operator Formulation}
Building upon this physical foundation, we propose the HCO as a differentiable module that embeds the heat diffusion equation into neural feature learning. Given two input modalities, $X_1$ (HSI) and $X_2$ (LiDAR), the features are first projected into a shared latent space through modality-specific Stem Layers:
\begin{equation}
    U_i = \text{BN}(\text{Conv}(\text{GELU}(\text{BN}(\text{Conv}(X_i))))), \quad i \in \{1,2\},
\end{equation}
resulting in $U_i \in \mathbb{R}^{P \times P \times 2C}$. To enable separate and joint modeling, each $U_i$ is divided along the channel dimension:
\begin{equation}
    [u_i^1, u_i^2] = \text{Split}(U_i), \quad u_i^j \in \mathbb{R}^{P \times P \times C},
\end{equation}
where $u_i^1$ is processed by modality-specific HCO Layers to extract discriminative spatial–spectral cues, while $u_i^2$ participates in shared HCO-based alignment.

The modality-specific and shared diffusions are formulated as:
\begin{equation}
\begin{aligned}
U_i^t &= \text{HCO}_i(u_i^{1},k_{q=i}), \quad i \in \{1,2\}, \\
U_3^t &= \text{HCO}(\text{Concat}(u_1^{2}, u_2^{2}),k_{q=3}),
\end{aligned}
\end{equation}
where $k_q$ denotes the effective diffusion coefficient used in the $q$-th HCO branch. To make the heat-conduction process learnable while preserving the physical constraint on diffusivity, we introduce a latent FVE $\theta_q$ and map it to $k_q$ through a learnable linear transformation followed by a ReLU activation:
\begin{equation}
    k_q = \operatorname{ReLU}(W_k\theta_q+b_k).
\end{equation}
Here, $\theta_q$ provides a flexible learnable representation for adapting the diffusion behavior to different modalities and scenes, while the non-negative mapping ensures that the effective diffusivity used in HCO remains consistent with the forward heat-conduction formulation. Depending on the FVE configuration, $k_q$ can further vary across spatial positions and channels, enabling adaptive diffusion for heterogeneous HSI and LiDAR features.

\subsubsection{HCO Architecture and Theoretical Properties}
Each HCO Layer adopts a “\text{LN–Operator–LN–FFN}” structure, inspired by implicit-state models such as Mamba and transformer-style visual operators, but with physically interpretable dynamics. Formally, for an input feature tensor $u \in \mathbb{R}^{P \times P \times C}$:
\begin{equation}
\begin{split}
    z_1 &= \text{LN}(u), \\
    z_2 &= \text{DWConv}(z_1), \\
    z_3 &= \sigma(\text{Linear}_1(z_2)) \odot \text{HCO}(\text{Linear}_2(z_2),k_q),
\end{split}
\end{equation}
where $\text{DWConv}$ captures local spatial dependencies and $\sigma(\cdot)$ denotes the SiLU activation. The two parallel linear projections serve as gating and conduction paths, respectively, with element-wise modulation $\odot$ implementing adaptive diffusion control.

The output is then normalized and updated through a residual and feed-forward structure:
\begin{equation}
    z_4 = \text{LN}(z_1 + z_3), \quad U_i^t = z_4 + \text{FFN}(z_4).
\end{equation}
This design allows the network to explicitly learn anisotropic, directionally varying diffusion patterns that reflect the geometry and texture structure of the input. Theoretically, $\mathrm{HCO}(\cdot,k_q)$ explicitly decouples physical diffusion from learnable refinement: the DCT-domain attenuation solves the discretized heat equation via Laplacian eigenmode modulation, while the auxiliary neural components function as data-adaptive calibration units.

\[
U^{(t+1)} = U^{(t)} + k \nabla^2 U^{(t)},
\]
where the Laplacian term $\nabla^2 U$ is implicitly realized in the frequency domain through DCT filtering. The learnable $k_q$ thus plays the role of a neural thermal diffusivity tensor, dynamically controlling the propagation of information across both space and channels. Unlike conventional gated blocks that rely entirely on data-driven routing, HCO anchors its computational core in an explicit heat-diffusion frequency response; the gating branch serves strictly to modulate this physical flow for heterogeneous HSI--LiDAR features.

By embedding this physically consistent process into multimodal representation learning, the HCO effectively bridges physics-based diffusion theory and deep spectral–spatial modeling. It achieves interpretable information propagation across modalities, providing enhanced robustness, generalization, and physical plausibility compared with purely data-driven fusion mechanisms.

\subsection{Multimodal Feature Fusion of M2Heat}

To capture complementary structures and semantic cues from hyperspectral and LiDAR modalities, M2Heat introduces a CFF module that explicitly operates in the frequency domain. This approach leverages both amplitude- and phase-based representations to achieve fine-grained cross-modal alignment and complementary feature enhancement.

\subsubsection{Frequency-domain decomposition}

Each input feature $U_i^t \in \mathbb{R}^{H \times W \times C}$ is first transformed into the frequency domain using the 2D Fast Fourier Transform (FFT):

\begin{equation}
    \mathcal{F}_i = \text{FFT}(U_i^t), \quad i \in \{1,2\}, \quad \mathcal{F}_i \in \mathbb{C}^{H \times W \times C}.
\end{equation}

We explicitly decompose $\mathcal{F}_i$ into amplitude and phase components:

\begin{equation}
    \mathcal{A}_i = |\mathcal{F}_i|, \quad 
    \mathcal{P}_i = \angle \mathcal{F}_i,
\end{equation}
where $\mathcal{A}_i \in \mathbb{R}^{H \times W \times C}$ encodes the energy distribution across spatial frequencies, and $\mathcal{P}_i \in [-\pi, \pi]^{H \times W \times C}$ preserves high-frequency structural details. This decomposition allows separate modeling of low-frequency semantic structures and high-frequency geometric details.

\begin{table*}[htbp]
\belowrulesep=0pt
\aboverulesep=0pt
\centering
\caption{Detailed Category Information for Three Datasets}

\small

\begin{tabular*}{\textwidth}{@{\extracolsep{\fill}}c||lcc||lcc||lcc}
\toprule
\multirow{2}{*}{No.} & \multicolumn{3}{c||}{Trento} & \multicolumn{3}{c||}{Houston2013} & \multicolumn{3}{c}{Augsburg} \\ \cline{2-10}
 & Class name & Train & Test & Class name & Train & Test & Class name & Train & Test \\ \midrule \midrule
C1  & Apple Trees          & 20 & 4014  & Healthy grass        & 198 & 1053 & Forest               & 146 & 13361 \\ 
C2  & Buildings            & 20 & 2883  & Stressed grass       & 190 & 1064 & Residential Area     & 264 & 30065 \\ 
C3  & Ground               & 20 & 459   & Synthetic grass      & 188 & 505  & Industrial Area      & 21  & 3830  \\ 
C4  & Woods                & 20 & 9103  & Trees                & 196 & 1072 & Low Plants            & 248 & 26609 \\ 
C5  & Vineyard             & 20 & 10481 & Soil                 & 186 & 1056 & Allotment             & 52  & 523   \\ 
C6  & Roads                & 20 & 3154  & Water                & 182 & 143  & Commercial Area       & 7   & 1638  \\ 
C7  &                      &    &       & Residential          & 196 & 1072 & Water                 & 23  & 1507  \\ 
C8  &                      &    &       & Commercial           & 191 & 1053 &                       &     &       \\  
C9  &                      &    &       & Road                 & 193 & 1059 &                       &     &       \\ 
C10 &                      &    &       & Highway              & 191 & 1036 &                       &     &       \\ 
C11 &                      &    &       & Railway              & 181 & 1054 &                       &     &       \\ 
C12 &                      &    &       & Parking Lot 1        & 192 & 1041 &                       &     &       \\ 
C13 &                      &    &       & Parking Lot 2        & 184 & 285  &                       &     &       \\ 
C14 &                      &    &       & Tennis Court         & 181 & 247  &                       &     &       \\ 
C15 &                      &    &       & Running Track        & 187 & 473  &                       &     &       \\ \midrule \midrule

-   & \textbf{Total}       & 120 & 30094 & \textbf{Total}       & 2832 & 12197 & \textbf{Total}       & 761 & 77533 \\ \bottomrule
\end{tabular*}
\label{tab:datasets}
\end{table*}

\subsubsection{Cross-modal frequency fusion}  
To align modalities in the spectral domain, the corresponding amplitude and phase components are concatenated and processed through a shared 2D MLP (implemented via convolution-GELU-convolution), enabling joint frequency-aware feature learning:

\begin{equation}
\begin{split}
    \hat{\mathcal{P}} &= \text{MLP}_{2D}\big(\text{Concat}(\mathcal{P}_1, \mathcal{P}_2)\big), \quad\\
    \hat{\mathcal{A}} &= \text{MLP}_{2D}\big(\text{Concat}(\mathcal{A}_1, \mathcal{A}_2)\big).
\end{split}
\end{equation}

This operation can be interpreted as learning a frequency-domain transfer function $T: (\mathcal{A}_1, \mathcal{A}_2, \mathcal{P}_1, \mathcal{P}_2) \mapsto (\hat{\mathcal{A}}, \hat{\mathcal{P}})$ that maximizes inter-modality mutual information while preserving intra-modality spectral energy.

\subsubsection{Spatial reconstruction}  
The fused frequency representations are transformed back to the spatial domain using the inverse FFT:

\begin{equation}
    \hat{U}_f^t = \text{IFFT}(\hat{\mathcal{A}} \odot e^{j \hat{\mathcal{P}}}), 
\end{equation}
where $\odot$ denotes element-wise multiplication and $j$ is the imaginary unit. This step reconstructs spatial features that integrate complementary low- and high-frequency information from both modalities.

\subsubsection{Weighted integration with HCO features}  
To exploit additional joint correlations, the frequency-fused feature $\hat{U}_f^t$ is combined with the HCO-aligned feature $U_3^t$ via learnable weights $\alpha$ and $\beta$:

\begin{equation}
    \hat{U}^t = \alpha \cdot \text{Conv}_{1 \times 1}(U_3^t) + \beta \cdot \text{Conv}_{1 \times 1}(\hat{U}_f^t),
\end{equation}
where the $1 \times 1$ convolution both aligns the channel dimension and introduces nonlinear refinement. The learnable parameters $\alpha, \beta \ge 0$ enforce an implicit energy-preserving constraint:

\begin{equation}
\begin{split}
\|\hat{U}^t\|_F^2 &\approx \alpha^2 \|\text{Conv}(U_3^t)\|_F^2 
+ \beta^2 \|\text{Conv}(\hat{U}_f^t)\|_F^2 \\
&\quad + 2 \alpha \beta \langle \text{Conv}(U_3^t), \text{Conv}(\hat{U}_f^t)\rangle,
\end{split}
\end{equation}
ensuring that the fused representation balances contributions from both spatial- and frequency-domain features.
 
The proposed CFF module constitutes a hybrid fusion mechanism, explicitly leveraging frequency-domain amplitude-phase decomposition and spatial-domain correlations. It should be noted that DCT and FFT serve different purposes in M2Heat. The HCO adopts DCT because its real-valued cosine basis naturally supports reflective boundary conditions and enables boundary-consistent heat diffusion over finite feature patches. In contrast, CFF employs FFT to obtain a complex-valued spectrum, allowing the amplitude and phase components to be explicitly separated and jointly interacted across modalities. By integrating cross-modal spectral interactions and energy-aware weighting, M2Heat achieves a theoretically grounded and empirically effective representation for joint classification tasks.

\begin{figure*}
    \centering
    \includegraphics[width=0.8\linewidth]{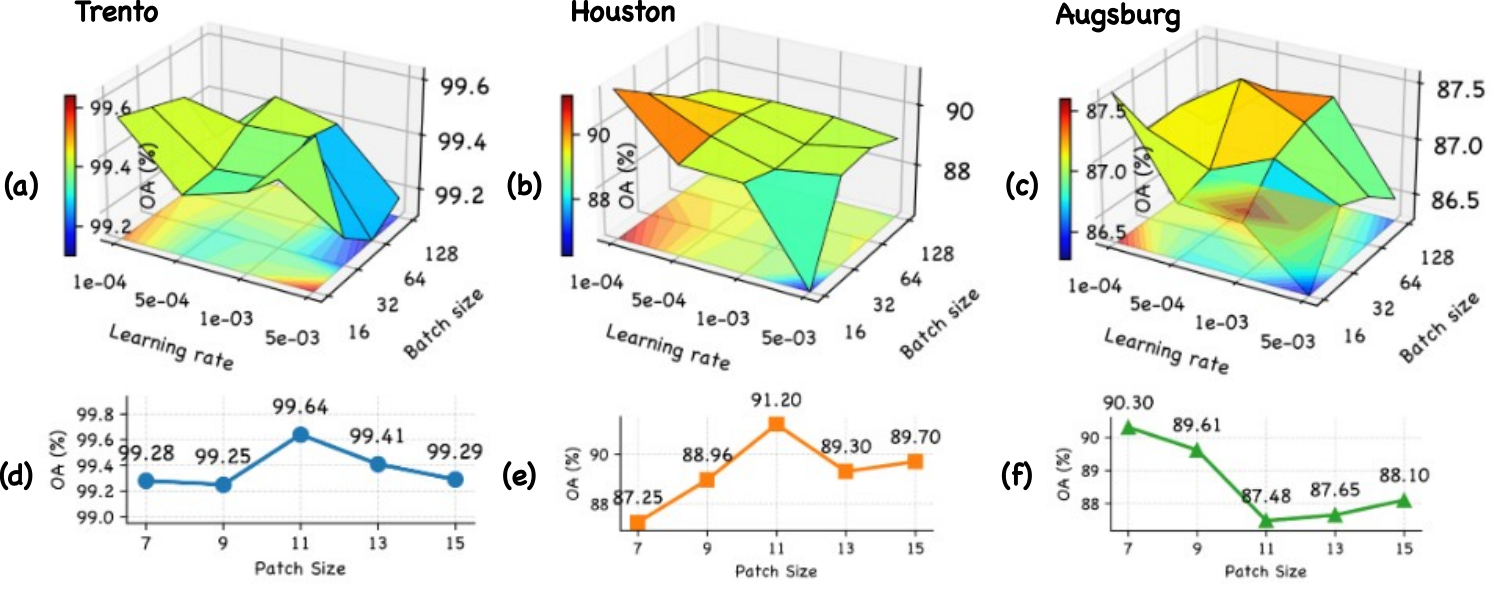}
    \caption{Impact of hyperparameter selection on the model. (a)--(c) OA variations under different learning rates and batch sizes on the Trento, Houston2013, and Augsburg datasets, respectively; (d)--(f) OA variations under different patch sizes.}
    \label{fig:fig5}
\end{figure*}

\subsection{Classification Mapping of M2Heat}

To generate semantic predictions from the fused features, M2Heat employs a lightweight yet effective classification head that leverages both spatially aligned and modality-aware information. Specifically, the fused feature $\hat{U}^t \in \mathbb{R}^{H \times W \times C}$ is first processed by a $1 \times 1$ convolution for channel reduction, followed by batch normalization (BN), LeakyReLU activation, and global average pooling (GAP) to capture global contextual information:

\begin{equation}
    z = \text{GAP}\Big(\phi\big(\text{BN}(\text{Conv}_{1\times 1}(\hat{U}^t))\big)\Big),
\end{equation}
where $\phi(\cdot)$ denotes the LeakyReLU activation function and $z \in \mathbb{R}^{C}$ is the compact feature vector for each input sample.

The resulting feature vector is then projected into the class space via a second $1 \times 1$ convolution followed by a softmax function to produce the normalized class probabilities:

\begin{equation}
    \tilde{Y} = \text{softmax}(\text{Conv}_{1\times 1}(z)) \in \mathbb{R}^{K},
\end{equation}
where $K$ denotes the number of classes.

For supervised training, we adopt the standard cross-entropy loss $\mathcal{L}_{\text{CE}}$ over all pixels in the training set:

\begin{equation}
    \mathcal{L}_{\text{CE}} = - \frac{1}{N} \sum_{i=1}^{N} \sum_{k=1}^{K} y_{i,k} \log(\tilde{y}_{i,k}),
\end{equation}
where $y_{i,k} \in \{0,1\}$ is the ground-truth label indicator for pixel $i$ and class $k$, $\tilde{y}_{i,k}$ is the predicted probability, and $N$ is the total number of pixels in the batch. This formulation enforces the network to maximize the likelihood of correct class assignments while maintaining the compact and modality-aware feature representation learned by M2Heat.

\begin{table*}[!htbp]
    \belowrulesep=0pt
    \aboverulesep=0pt
    \centering
    \caption{Classification accuracy (\%) obtained by different methods on the Trento dataset.}
    \begin{tabular*}{\textwidth}{@{\extracolsep{\fill}}@{}c||cc||ccccccc||c@{}}
    \toprule
         No. &  SF& MambaHSI& MFT& HCT& DSHFNet& ExViT& Cross-HL& HLMamba& S2CMamba& M2Heat\\
         Year&  2022~\cite{hongSpectralFormerRethinkingHyperspectral2022}& 2024~\cite{mambahsi}& 2023~\cite{royMultimodalFusionTransformer2023a}& 2023~\cite{zhaoJointClassificationHyperspectral2023}& 2023~\cite{fengDSHFNetDynamicScale2023}& 2023~\cite{yaoExtendedVisionTransformer2023}& 2024~\cite{royCrossHyperspectralLiDAR2024a}& 2024~\cite{liaoJointClassificationHyperspectral2024}& 2024~\cite{zhangS2CrossMambaSpatialSpectral2024}& --\\
        \midrule
        \midrule
         1& 91.75& 99.28& 98.03& 98.93& 99.05& 99.48& \textbf{100}& 99.05& 98.28& 99.88\\
         2& 82.83& 85.81& 99.03& 97.85& 96.50& \textbf{99.20}& 98.99& 97.50& 99.10& 98.75\\
         3& 94.77& 94.55& 99.56& \textbf{100}& 97.17& \textbf{100}& 99.35& 99.13& \textbf{100}& 99.56\\
         4& 99.51& 99.99& \textbf{100}& \textbf{100}& 99.96& \textbf{100}& \textbf{100}& \textbf{100}& \textbf{100}& \textbf{100}\\
         5& 99.17& 98.36& \textbf{100}& \textbf{100}& \textbf{100}& 99.99& 99.96& 99.86& \textbf{100}& \textbf{100}\\
         6& 78.79& 82.56& 94.99& 93.02& 93.47& 94.61& 86.24& 89.06& 95.43& \textbf{97.91}\\
         \midrule
         \midrule
         OA(\%)& 94.51& 96.06& 99.11& 98.92& 98.80& 99.29& 98.44& 98.42& 99.21& \textbf{99.64}\\
         AA(\%)& 91.14& 93.43& 98.60& 98.30& 97.69& 98.88& 97.42& 97.43& 98.80& \textbf{99.35}\\
         $\kappa$& 0.9267& 0.9475& 0.9882& 0.9856& 0.9839& 0.9905& 0.9792& 0.9790& 0.9894& \textbf{0.9952}\\
         \bottomrule 
    \end{tabular*}
    \label{tab:trento}
\end{table*}

\begin{figure*}
    \centering
    \includegraphics[width=0.95\linewidth]{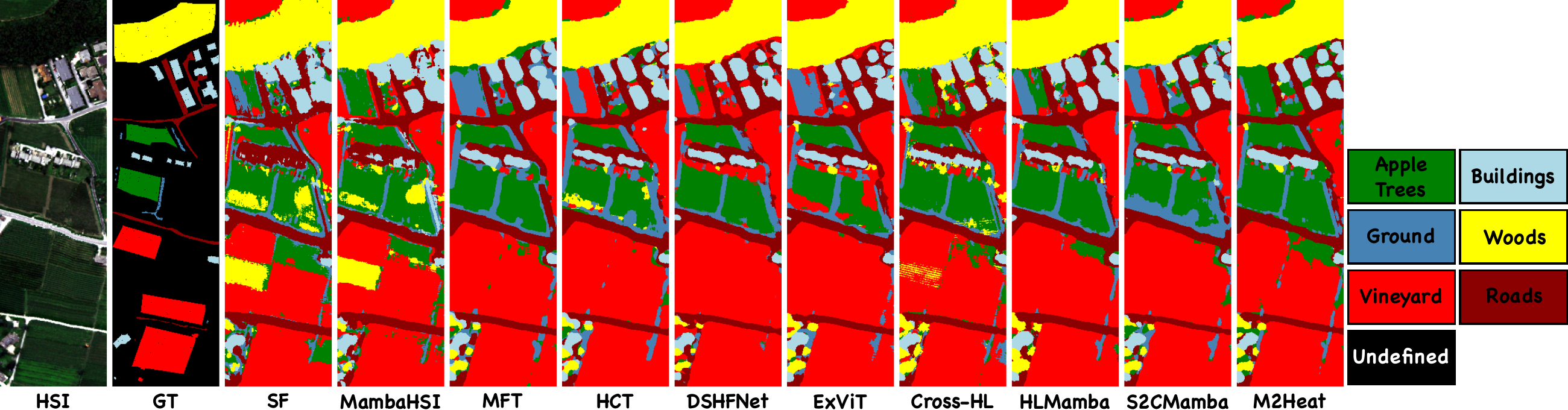}
    \caption{Classification result mapping by different models on the Trento dataset.}
    \label{fig:fig6}
\end{figure*}

\section{EXPERIMENT AND DISCUSSION}
\label{sec:exp}
This section presents a comprehensive experimental evaluation of the proposed M2Heat model, employing three prominent HSI-LiDAR benchmark datasets: Trento, Houston2013, and Augsburg. We begin by delineating the characteristics of each dataset, followed by a detailed description of our experimental protocol and the adopted evaluation metrics. To rigorously assess its performance, M2Heat is benchmarked against nine state-of-the-art methods. Finally, to substantiate the efficacy of our design choices and demonstrate the model's robustness, we conduct extensive ablation studies and supplementary analyses across all three datasets.

\subsection{Datasets Description}

\subsubsection{Trento} 
The Trento dataset was acquired over a rural area near Trento, located in the southern part of Italy. It comprises co-registered HSI and LiDAR data. The scene has a spatial size of 166 × 600 pixels with a ground sampling distance of 1 m. The hyperspectral data cover the spectral range from 0.42 to 0.99 $\mu m$, comprising 63 contiguous bands after removing noisy channels. The LiDAR measurements were collected using an Optech ALTM 3100EA sensor, providing elevation information complementary to the spectral content. The dataset contains a total of 30,214 labeled samples distributed over six land use classes, making it a widely used benchmark for multimodal RS classification.

\subsubsection{Houston2013} 
\footnote{\url{http://www.classic.grss-ieee.org/community/technical-committees/datafusion/2013-ieee-grss-data-fusion-contest/}}
The Houston2013 dataset was collected over the University of Houston campus and its neighboring urban areas during the 2013 IEEE GRSS Data Fusion Contest. The HSI was acquired using the Compact Airborne Spectral Imager (CASI), with a spatial size of 345 × 1905 pixels and a ground sampling distance of 2.5 m. The spectral range spans 0.38–1.35 $\mu m$, consisting of 144 contiguous bands after removing water absorption and noisy channels. In addition, LiDAR-derived Digital Surface Model (DSM) data were acquired over the same area to provide complementary elevation information. The dataset contains 15,029 labeled samples distributed across 15 land-cover classes, offering diverse urban and vegetation categories for evaluating multimodal data fusion methods.

\begin{table*}[!htbp]
    \belowrulesep=0pt
    \aboverulesep=0pt
    \centering
    \caption{Classification accuracy (\%) obtained by different methods on the Houston2013 dataset.}
    \begin{tabular*}{\textwidth}{@{\extracolsep{\fill}}@{}c||cc||ccccccc||c@{}}
    \toprule
         No. &  SF& MambaHSI& MFT& HCT& DSHFNet& ExViT& Cross-HL& HLMamba& S2CMamba& M2Heat\\
         Year&  2022~\cite{hongSpectralFormerRethinkingHyperspectral2022}& 2024~\cite{mambahsi}& 2023~\cite{royMultimodalFusionTransformer2023a}& 2023~\cite{zhaoJointClassificationHyperspectral2023}& 2023~\cite{fengDSHFNetDynamicScale2023}& 2023~\cite{yaoExtendedVisionTransformer2023}& 2024~\cite{royCrossHyperspectralLiDAR2024a}& 2024~\cite{liaoJointClassificationHyperspectral2024}& 2024~\cite{zhangS2CrossMambaSpatialSpectral2024}& --\\
        \midrule
        \midrule
         1& 81.29& 80.44& 79.68& 82.15& 82.72& 82.72& \textbf{83.10}& 83.00& 81.77& 81.39\\
         2& 75.85& 84.68& \textbf{96.24}& 84.77& 84.21& 84.59& 84.87& 85.15& 83.55& 84.96\\
         3& 58.61& 54.46& 96.63& 97.31& 97.03& 98.41& \textbf{98.42}& 88.32& 97.82& 96.24\\
         4& 85.61& 88.73& 95.55& 97.82& 89.68& 92.71& 93.09& \textbf{100}& 89.77& 90.53\\
         5& 98.96& 97.44& \textbf{100}& 99.91& 99.91& \textbf{100}& 99.91& \textbf{100}& \textbf{100}& \textbf{100}\\
         6& 95.80& \textbf{100}& 79.02& 95.80& 98.60& 99.30& 95.80& \textbf{100}& 81.82& \textbf{100}\\
         7& 84.79& 81.90& 88.15& 92.78& 74.25& 92.68& 86.19& 87.78& 90.02& \textbf{92.82}\\
         8& 52.90& 64.48& 83.84& 92.78& 73.03& 85.16& 83.10& 77.49& 88.70& \textbf{95.25}\\
         9& 76.96& 77.53& 81.40& 85.55& 73.09& 84.99& 83.29& 84.42& 83.66& \textbf{86.97}\\
         10& 51.16& 50.97& 59.17& 65.83& 67.86& 65.57& 51.93& 61.29& 57.53& \textbf{79.25}\\
         11& 54.74& 57.69& 81.69& 94.02& 71.73& 84.25& 82.64& 79.51& 93.07& \textbf{99.43}\\
         12& 83.57& 80.31& 88.95& 86.36& 92.41& 90.87& 92.03& 87.03& \textbf{96.16}& 93.47\\
         13& 66.67& 87.37& 84.56& 88.42& 56.49& 91.93& 91.23& 90.53& \textbf{92.98}& 84.56\\
         14& 91.90& 80.57& \textbf{100}& 93.93& 97.17& 99.19& 93.12& 98.79& 99.60& \textbf{100}\\
         15& 69.98& 69.13& 98.52& 83.93& 76.53& 96.41& \textbf{100}& 92.18& 99.37& \textbf{100}\\
         \midrule
         \midrule
         OA(\%)& 74.21& 75.90& 87.03& 88.92& 81.36& 89.07& 85.77& 85.06& 87.80& \textbf{91.20}\\
         AA(\%)& 75.25& 77.05& 87.56& 89.43& 82.31& 89.64& 87.91& 87.23& 89.05& \textbf{92.32}\\
         $\kappa$& 0.7214& 0.7397& 0.8595& 0.8800& 0.7982& 0.8817& 0.8461& 0.8385& 0.8678& \textbf{0.9045}\\
         \bottomrule 
    \end{tabular*}
    \label{tab:houston}
\end{table*}

\begin{figure*}
    \centering
    \includegraphics[width=0.95\linewidth]{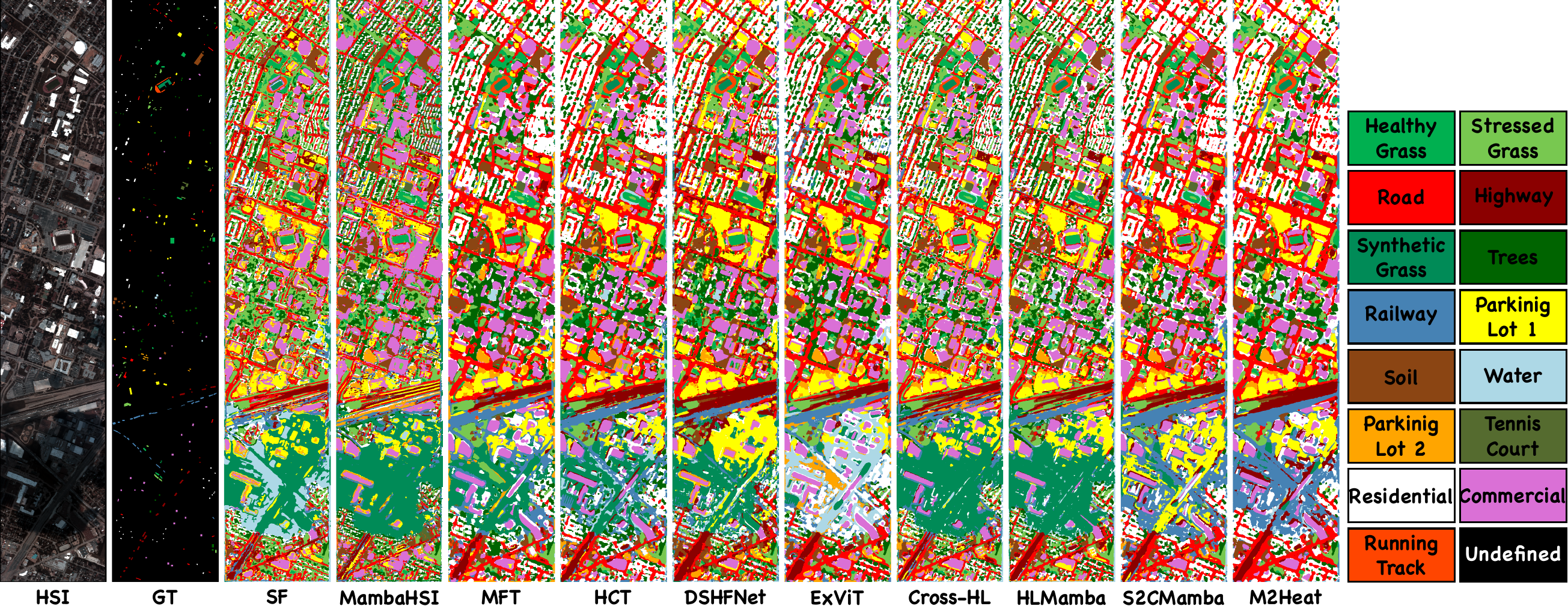}
    \caption{Classification result mapping by different models on the Houston dataset.}
    \label{fig:fig7}
\end{figure*}

\subsubsection{Augsburg~\cite{hongMultimodalRemoteSensing2021a}} 
\footnote{\url{https://github.com/danfenghong/ISPRS_S2FL}}
The Augsburg dataset covers an urban area in the vicinity of Augsburg, Germany, and includes a spaceborne hyperspectral (HSI) image and a digital surface model (DSM) derived from LiDAR measurements. The hyperspectral data were acquired by the HySpex sensor, providing 180 spectral bands spanning the 0.4–2.5~$\mu$m wavelength range. The DSM data were captured by the DLR-3K LiDAR system, providing precise elevation information for the same area. For consistency and computational efficiency, all modalities were resampled to a unified spatial resolution of 30~m ground sampling distance (GSD), resulting in an image of size $332 \times 485$ pixels. Each pixel is thus associated with a 180-dimensional spectral vector from the HSI and a corresponding elevation value from the DSM. Ground-truth labels were derived from OpenStreetMap, covering multiple land-cover categories suitable for multimodal classification and fusion research, including urban, vegetation, water, and impervious surfaces. 

The Houston2013 and Augsburg datasets provide publicly available ground-truth annotations with standardized training–testing splits, which were directly adopted in our experiments to ensure comparability with existing works. In contrast, the Trento dataset does not include an official split. Following a class-balanced sampling strategy, we randomly selected 20 labeled pixels per class for training, with the remaining samples used for testing. Detailed category information for the three datasets, along with the division of training and test sets, is shown in Table~\ref{tab:datasets}.

\subsection{Experimental Setup}

All experiments in this study were implemented using PyTorch 2.0 and conducted on a Linux workstation equipped with an Intel(R) Xeon(R) Gold 6133 CPU @ 2.50 GHz and an NVIDIA GeForce RTX 4090 GPU with CUDA 11.8 support. The M2Heat model is optimized using the Adam optimizer with a weight decay of 0 and trained for 500 epochs. The learning rate was scheduled using StepLR with a step size of 20 and a decay factor of 0.9. The hyperspectral data were normalized along the spectral dimension before training. Dataset-specific learning rates, batch sizes, and patch sizes are reported in the subsequent hyperparameter selection section. For SOTA comparison methods, we strictly adhered to the hyperparameter settings and training protocols reported in their original publications. In cases where such information was unavailable, we optimized the hyperparameters under the same training regime as our method to ensure a fair and consistent comparison.

To comprehensively evaluate classification performance, three widely accepted metrics were employed: Overall Accuracy (OA), Average Accuracy (AA), and the Kappa coefficient ($\kappa$).

\renewcommand{\arraystretch}{1.08}
\begin{table*}[!htbp]
    \belowrulesep=0pt
    \aboverulesep=0pt
    \centering
    \caption{Classification accuracy (\%) obtained by different methods on the Augsburg dataset.}
    \resizebox{\linewidth}{!}{
    \begin{tabular*}{\textwidth}{@{\extracolsep{\fill}}@{}c||cc||ccccccc||c@{}}
    \toprule
         No. &  SF& MambaHSI& MFT& HCT& DSHFNet& ExViT& Cross-HL& HLMamba& S2CMamba& M2Heat\\
         Year&  2022~\cite{hongSpectralFormerRethinkingHyperspectral2022}& 2024~\cite{mambahsi}& 2023~\cite{royMultimodalFusionTransformer2023a}& 2023~\cite{zhaoJointClassificationHyperspectral2023}& 2023~\cite{fengDSHFNetDynamicScale2023}& 2023~\cite{yaoExtendedVisionTransformer2023}& 2024~\cite{royCrossHyperspectralLiDAR2024a}& 2024~\cite{liaoJointClassificationHyperspectral2024}& 2024~\cite{zhangS2CrossMambaSpatialSpectral2024}& --\\
        \midrule
        \midrule
         1& 80.08& 90.14& 91.62& 92.96& \textbf{97.15}& 92.52& 95.77& 96.70& 92.63& 95.61\\
         2& 87.74& 94.59& 94.91& 96.46& 97.95& 95.58& 94.93& 96.76& 97.25& 96.31\\
         3& 71.12& 2.01& \textbf{68.46}& 54.76& 29.97& 44.07& 44.49& 58.43& 43.79& 66.08\\
         4& 78.62& 84.06& 86.24& 85.97& 84.31& 86.45& 85.55& 82.28& 87.15& \textbf{92.14}\\
         5& 34.03& 39.77& 41.32& 40.36& 39.48& 41.93& 38.27& 41.93& 42.08& \textbf{61.76}\\
         6& 7.94& 0.79& 28.56& 23.44& 8.32& \textbf{36.66}& 25.75& 13.55& 24.27& 9.95\\
         7& 11.68& 15.93& 34.78& 35.21& 29.93& 33.75& 28.04& 27.27& 32.45& \textbf{49.44}\\
         \midrule
         \midrule
         OA(\%)& 78.94& 81.75& 84.46& 86.47& 86.04& 87.18& 86.28& 87.20& 87.11& \textbf{90.30}\\
         AA(\%)& 53.03& 46.75& \textbf{63.70}& 61.30& 55.30& 61.71& 60.40& 59.56& 59.94& \textbf{67.33}\\
         $\kappa$& 0.6984& 0.7318& 0.7821& 0.8162& 0.7972& 0.8164& 0.8030& 0.8162& 0.8153& \textbf{0.8607}\\
         \bottomrule 
    \end{tabular*}
    }
    \label{tab:augsburg}
\end{table*}

\begin{figure*}
    \centering
    \includegraphics[width=0.95\linewidth]{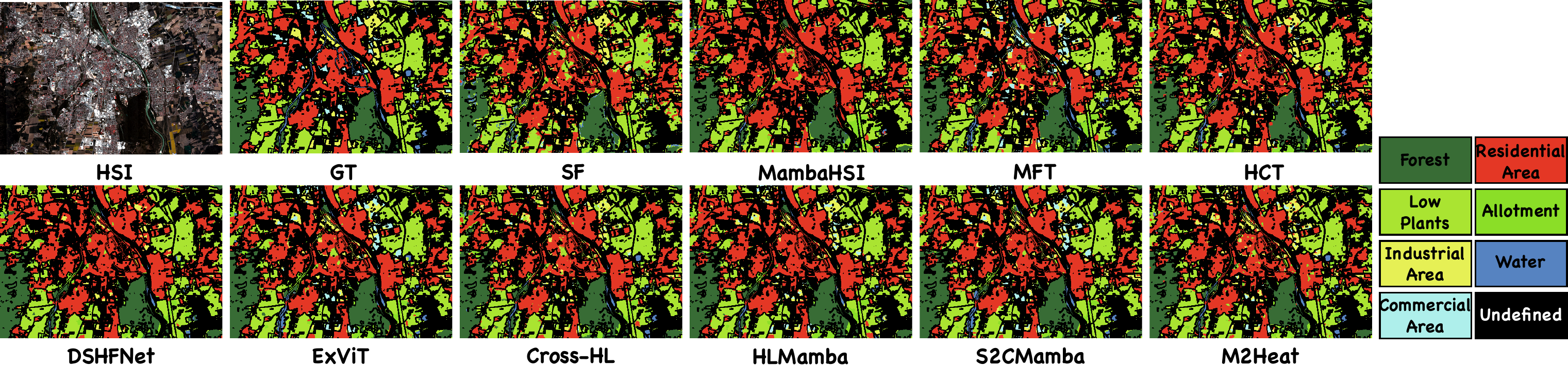}
    \caption{Classification result mapping by different models on the Augsburg dataset.}
    \label{fig:fig8}
\end{figure*}

\subsection{Hyperparameter Selection}

In deep learning-based classification tasks, the choice of training hyperparameters, particularly the learning rate and batch size, plays a critical role in model convergence and generalization performance.

To identify the optimal learning rate and batch size for the proposed M2Heat model, we conducted a series of grid search experiments on three benchmark datasets: Trento, Houston2013, and Augsburg. The candidate learning rates were set to \{1e-4, 5e-4, 1e-3, 5e-3\}, and batch sizes were selected from \{16, 32, 64, 128\}. As shown in Fig.~\ref{fig:fig5}, the performance was evaluated using Overall Accuracy (OA) on the validation sets.

For the Trento dataset, the highest OA of 99.64\% was achieved with a learning rate of 5e-3 and a batch size of 16. Notably, smaller batch sizes tended to yield slightly better performance, while variations in learning rate exhibited relatively minor influence on accuracy within the tested range.

On the Houston2013 dataset, the optimal configuration was a learning rate of 1e-4 combined with a batch size of 16, which attained an OA of 91.2\%. The results indicated that lower learning rates generally facilitated more stable convergence, while increasing batch size did not significantly improve accuracy.

For the Augsburg dataset, the highest OA of 87.48\% was obtained with a learning rate of 5e-4 and a batch size of 64. Unlike the other datasets, moderate batch sizes coupled with mid-range learning rates yielded better performance, suggesting a more balanced trade-off between training stability and convergence speed.

To further examine the influence of spatial context on the heat-diffusion process, we evaluated M2Heat with patch sizes of ${7,9,11,13,15}$ while fixing the learning rate and batch size to the dataset-specific settings identified above. As illustrated in Fig.~\ref{fig:fig4}(d), M2Heat exhibits stable performance over a broad range of spatial neighborhoods, although the optimal patch size varies across datasets. Trento achieves its highest OA of $99.64\%$ with an $11\times11$ patch, while Houston2013 similarly peaks at $91.20\%$ under the same setting. For Augsburg, a smaller $7\times7$ patch yields the best OA of $90.30\%$, whereas larger patches provide lower performance. This result suggests that increasing the spatial context does not necessarily lead to monotonic improvements, as overly large neighborhoods may introduce heterogeneous or class-irrelevant information. The dataset-dependent optima therefore reflect differences in spatial resolution, scene composition, and local class distributions.

Based on these findings, the M2Heat model adopts dataset-specific hyperparameters for optimal performance: a learning rate of 5e-3, batch size of 16, and a patch size of $11\times11$ for Trento, 1e-4, 16, and $11\times11$ for Houston2013, and 5e-4, 64, and $7\times7$ for Augsburg. This tailored hyperparameter selection ensures robust and efficient training across diverse hyperspectral and LiDAR fusion scenarios. For a fair and controlled comparison in the ablation studies, the patch size is fixed to $11\times11$ for all datasets, so that the observed performance differences mainly reflect the effects of the investigated components rather than variations in the input neighborhood.

\subsection{Comparison Result and Analysis}

To comprehensively evaluate the effectiveness of the proposed M2Heat model, we compare it against nine SOTA HSI classification methods. These include two single-source HSI classification approaches, namely SpectralFormer (SF)~\cite{hongSpectralFormerRethinkingHyperspectral2022} and MambaHSI~\cite{mambahsi} which are trained on HSI only, as well as seven multisource joint classification methods integrating HSI and LiDAR data: MFT~\cite{royMultimodalFusionTransformer2023a}, HCT~\cite{zhaoJointClassificationHyperspectral2023}, DSHFNet~\cite{fengDSHFNetDynamicScale2023}, ExViT~\cite{yaoExtendedVisionTransformer2023}, Cross-HL~\cite{royCrossHyperspectralLiDAR2024a}, HLMamba~\cite{liaoJointClassificationHyperspectral2024}, and S$^2$CrossMamba (S2CMamba)~\cite{zhangS2CrossMambaSpatialSpectral2024}.

\subsubsection{Trento Dataset} 

We first present the classification performance of the competing methods on the Trento dataset, which contains complex urban and natural land cover classes captured by both HSI and LiDAR modalities. The quantitative results, including OA, AA, $\kappa$, and classwise accuracy, are summarized in Table~\ref{tab:trento}. The corresponding classification maps with a unified color legend are illustrated in Fig.~\ref{fig:fig6}.

As observed, conventional single-source HSI methods such as SF and MambaHSI achieve relatively lower OA values of 94.51\% and 96.06\%, respectively, reflecting the limitation of relying solely on spectral information. Multisource fusion methods generally outperform these baselines by leveraging complementary spatial and elevation cues.

Among multisource methods, MFT, HCT, and DSHFNet demonstrate strong performances with OA above 98.8\%, yet still fall short compared to transformer-based and hybrid models. ExViT and Cross-HL achieve notably high accuracies, with Cross-HL reaching perfect classification on four out of six classes and an overall OA of 98.44\%. The HLMamba and S2CMamba methods further improve the performance, with OA values of 98.42\% and 99.21\%, respectively, benefiting from enhanced feature extraction and fusion strategies. The proposed M2Heat model attains the highest OA of 99.64\%, surpassing all competitors by a clear margin. It achieves the best or near-best classification accuracies across all classes, especially excelling in the challenging sixth class with 97.91\% accuracy. Correspondingly, the $\kappa$ reaches 0.9952, indicating near-perfect agreement.

Visually, the classification maps in Fig.~\ref{fig:fig6} reveal that M2Heat produces spatially coherent and sharply delineated land cover regions. Compared to other methods, M2Heat better preserves fine structural details and boundaries, reducing misclassification in mixed or transitional areas. The overall classification maps generated by competing methods exhibit some degree of noise and misclassification patches, particularly in classes with fewer subtle spectral differences.

In summary, both quantitative and qualitative results validate the effectiveness of M2Heat in exploiting the complementary information of HSI and LiDAR data for accurate land cover classification on the Trento dataset.

\renewcommand{\arraystretch}{1.08}
\begin{table}
\belowrulesep=0pt
\aboverulesep=0pt
\centering
\caption{Overall accuracy (OA \%) of the ablation study.}
\label{tab:ablation}

\footnotesize
\setlength{\tabcolsep}{5pt}

\begin{tabular}{ccccc||ccc}
\toprule
\multirow{2}{*}{HSI} & \multirow{2}{*}{LiDAR} & \multirow{2}{*}{Joint} & \multicolumn{2}{c||}{CCF} & \multicolumn{3}{c}{OA (\%)} \\
\cmidrule(lr){4-5} \cmidrule(lr){6-8}
&  &  & S1 & S2 & Trento & Houston & Augsburg \\
\midrule \midrule
\checkmark &  &  &  &  & 97.63 & 86.55 & 83.69 \\
 & \checkmark &  &  &  & 97.92 & 60.31 & 76.98 \\
 &  & \checkmark &  &  & 99.18 & 89.80 & 85.90 \\
\checkmark & \checkmark &  & \checkmark &  & 99.26 & 87.82 & 84.95 \\
\checkmark &  & \checkmark & \checkmark &  & 99.23 & 90.05 & 87.23 \\
 & \checkmark & \checkmark &  & \checkmark & 99.26 & 88.61 & 86.57 \\
\checkmark & \checkmark & \checkmark & \checkmark & \checkmark & \textbf{99.64} & \textbf{91.20} & \textbf{87.48} \\
\bottomrule
\end{tabular}
\end{table}

\begin{table}[t]
\belowrulesep=0pt
\aboverulesep=0pt
\centering
\caption{OA (\%) comparison of different frequency representations in CFF and operators in HCO.}
\label{tab:transform_ablation}

\footnotesize
\setlength{\tabcolsep}{2.5pt}
\renewcommand{\arraystretch}{1.08}

\begin{tabular}{c|cccc|cccc}
\toprule
\multirow{2}{*}{Dataset}
& \multicolumn{4}{c|}{CFF}
& \multicolumn{4}{c}{HCO} \\
\cmidrule(lr){2-5}
\cmidrule(lr){6-9}
& DCT & FFT-RI & \textbf{FFT-AP} & FFT-L/H
& FFT & DST & \textbf{DCT} & GFNet \\
\midrule \midrule 
Trento
& 99.51 & 99.27 & \textbf{99.64 }& 99.44
& 99.24 & 99.17 & \textbf{99.64} & 99.56 \\

Houston
& 89.96 & 90.68 & \textbf{91.20} & 90.10
& \textbf{91.47} & 90.38 & 91.20 & 90.66 \\

Augsburg
& 87.06 & 86.23 & \textbf{87.48}& 86.98
& 87.07 & 86.28 & \textbf{87.48} & 85.43 \\
\bottomrule
\end{tabular}
\end{table}

\begin{table}[t]
\centering
\belowrulesep=0pt
\aboverulesep=0pt
\caption{Operator-level comparison of different modules. T: Trento; H: Houston2013; A: Augsburg; P: Parameters; F: FLOPs; M: GPU memory; I: inference time; $C_{\mathrm{t}}$: operator-level theoretical complexity. M and I are measured on the Trento dataset under the same hardware and evaluation protocol.}
\label{tab:operator_analysis}

\footnotesize
\setlength{\tabcolsep}{2.5pt}
\renewcommand{\arraystretch}{1.08}

\begin{tabular}{@{}c|ccc|cc|cc|c@{}}
\toprule
\multirow{2}{*}{Operators}
& \multicolumn{3}{c|}{OA (\%)}
& \multicolumn{2}{c|}{Model Statistics}
& \multicolumn{2}{c|}{Runtime Metrics}
& \multirow{2}{*}{$C_{\mathrm{t}}$} \\
\cmidrule(lr){2-4}
\cmidrule(lr){5-6}
\cmidrule(lr){7-8}
& T & H & A
& P (K) & F (G)
& M (MB) & I (s)
& \\
\midrule \midrule
Conv
& 99.26 & 88.92 & 85.60
& 793.71 & 8.444
& \textbf{425.84} & 0.62
& $\mathcal{O}(N)$ \\

MSA
& 99.23 & 87.78 & 87.10
& 588.41 & 5.632
& 539.83 & 0.68
& $\mathcal{O}(N^2)$ \\

Mamba
& 99.22 & 88.14 & 87.21
& 593.16 & \textbf{5.189}
& 704.63 & 0.75
& $\mathcal{O}(N)$ \\

HCO
& \textbf{99.64} & \textbf{91.20} & \textbf{87.48}
& \textbf{555.62} & 5.334
& 490.72 & \textbf{0.53}
& $\mathcal{O}(N^{1.5})$ \\
\bottomrule
\end{tabular}
\end{table}

\begin{table}[t]
\belowrulesep=0pt
\aboverulesep=0pt
\centering
\caption{OA (\%) and relative OA decay (ROD, \%) under different training-sample ratios. ROD is computed relative to the performance using 100\% training samples, where a lower value indicates better robustness.}
\label{tab:sample_robustness}

\footnotesize
\setlength{\tabcolsep}{3pt}
\renewcommand{\arraystretch}{1.08}

\begin{tabular}{@{}c|c|ccc|cccc@{}}
\toprule
\multirow{2}{*}{Dataset}
& \multirow{2}{*}{Operators}
& \multicolumn{3}{c|}{OA (\%) $\uparrow$}
& \multicolumn{4}{c}{ROD (\%) $\downarrow$} \\
\cmidrule(lr){3-5}
\cmidrule(lr){6-9}
& & 25\% & 50\% & 75\%
& 25\% & 50\% & 75\% & Avg. \\
\midrule \midrule

\multirow{4}{*}{Trento}
& Conv   & 98.68 & 99.12 & 99.15
& 0.580 & 0.146 & 0.112 & 0.279 \\
& MSA   & 97.56 & 97.97 & 98.29
& 1.679 & 1.270 & 0.949 & 1.299 \\
& Mamba & 97.84 & 99.20 & 98.66
& 1.394 & \textbf{0.024} & 0.567 & 0.662 \\
& HCO  & \textbf{99.18} & \textbf{99.45} & \textbf{99.54}
& \textbf{0.465} & 0.191 & \textbf{0.102} & \textbf{0.252} \\
\midrule

\multirow{4}{*}{Houston}
& Conv   & 87.62 & 87.69 & 87.98
& 1.462 & 1.379 & 1.056 & 1.299 \\
& MSA   & 86.24 & 86.63 & 87.56
& 1.751 & 1.312 & \textbf{0.246} & 1.103 \\
& Mamba & 86.95 & 87.26 & 87.77
& 1.347 & 0.999 & 0.422 & 0.923 \\
& HCO  & \textbf{90.32} & \textbf{90.56} & \textbf{90.69}
& \textbf{0.964} & \textbf{0.698} & 0.563 & \textbf{0.742} \\
\midrule

\multirow{4}{*}{Augsburg}
& Conv   & 84.60 & 85.32 & 85.10
& 1.165 & \textbf{0.333} & 0.583 & 0.694 \\
& MSA   & 85.91 & 85.76 & 86.12
& 1.369 & 1.542 & 1.123 & 1.345 \\
& Mamba & 86.10 & 86.26 & 85.74
& 1.273 & 1.085 & 1.690 & 1.350 \\
& HCO  & \textbf{86.80} & \textbf{87.10} & \textbf{87.26}
& \textbf{0.779} & 0.438 & \textbf{0.254} & \textbf{0.490} \\
\bottomrule
\end{tabular}
\end{table}

\subsubsection{Houston2013 Dataset}

We further evaluate the classification performance of all competing methods on the Houston2013 dataset, which comprises diverse urban land cover classes characterized by complex spatial structures and spectral variability. Table \ref{tab:houston} reports the quantitative results, while Fig.~\ref{fig:fig7} presents the corresponding classification maps with a unified legend.

As shown, single-source HSI methods SF and MambaHSI yield relatively modest OA values of 74.21\% and 75.90\%, respectively, highlighting the limitations of using spectral information alone in this challenging scenario. Among multisource fusion techniques, CNN-based methods such as MFT and HCT achieve significantly improved results, with OA surpassing 87\%, reflecting the benefit of integrating LiDAR elevation data. Transformer-based methods like ExViT and Cross-HL demonstrate competitive performances, achieving OA values of 89.07\% and 85.77\%, respectively. However, the recently proposed HLMamba and S2CMamba methods yield slightly lower OA values of 85.06\% and 87.80\%, suggesting room for improvement in their fusion strategies.Our proposed M2Heat model achieves favorable overall performance, attaining an OA of 91.20\%, AA of 92.32\%, and the highest Kappa coefficient of 0.9045. Notably, M2Heat achieves the best classification accuracies across several challenging classes, including classes Residential, Commercial, Highway, and Railway, which correspond to complex urban features with subtle spectral and spatial differences.

Visual inspection of the classification maps (Fig.~\ref{fig:fig7}) reveals that M2Heat better handles challenging regions affected by shadows and mixed pixels, which are common in hyperspectral urban scenes. In particular, the shadowed area caused by vegetation and tall buildings, which commonly degrades spectral quality, is more accurately classified by M2Heat compared to other methods. Additionally, several competing models exhibit evident misclassifications between Water and Railway classes, likely due to spectral confusion and limited spatial context modeling. In contrast, M2Heat delineates these classes with clearer boundaries and fewer errors, demonstrating its ability to exploit complementary spectral, spatial, and elevation information.

\subsubsection{Augsburg Dataset}

The classification results on the Augsburg dataset are summarized in Table \ref{tab:augsburg}, with the corresponding maps shown in Fig.~\ref{fig:fig8}. Single-source HSI methods SF and MambaHSI achieve lower overall accuracies of 78.94\% and 81.75\%, respectively, demonstrating the challenges posed by this dataset’s diverse land cover types.

Among multisource methods, MFT, HCT, DSHFNet, ExViT, Cross-HL, HLMamba, and S2CMamba exhibit comparable performances, with OAs clustered around 86–87\%. Our proposed M2Heat achieves the highest OA of 90.30\% and the highest Kappa coefficient of 0.8607, confirming its robustness in integrating spectral and elevation information.

While M2Heat shows marginal gains in overall metrics, it notably improves classification accuracies in challenging classes such as classes Residential Area, Low Plants, Allotment, and Water, which correspond to heterogeneous urban features with subtle spectral differences. However, all methods struggle with classes Industrial Area and Commercial Area due to limited training samples and class imbalance, which is a known difficulty in this dataset.

Visual comparison reveals that M2Heat produces smoother and more spatially consistent classification maps, effectively reducing noise and misclassification in complex regions. This result indicates that M2Heat can effectively exploit complementary information from HSI and LiDAR data for fine-grained urban land cover classification.

\begin{figure}
    \centering
    \includegraphics[width=0.95\linewidth]{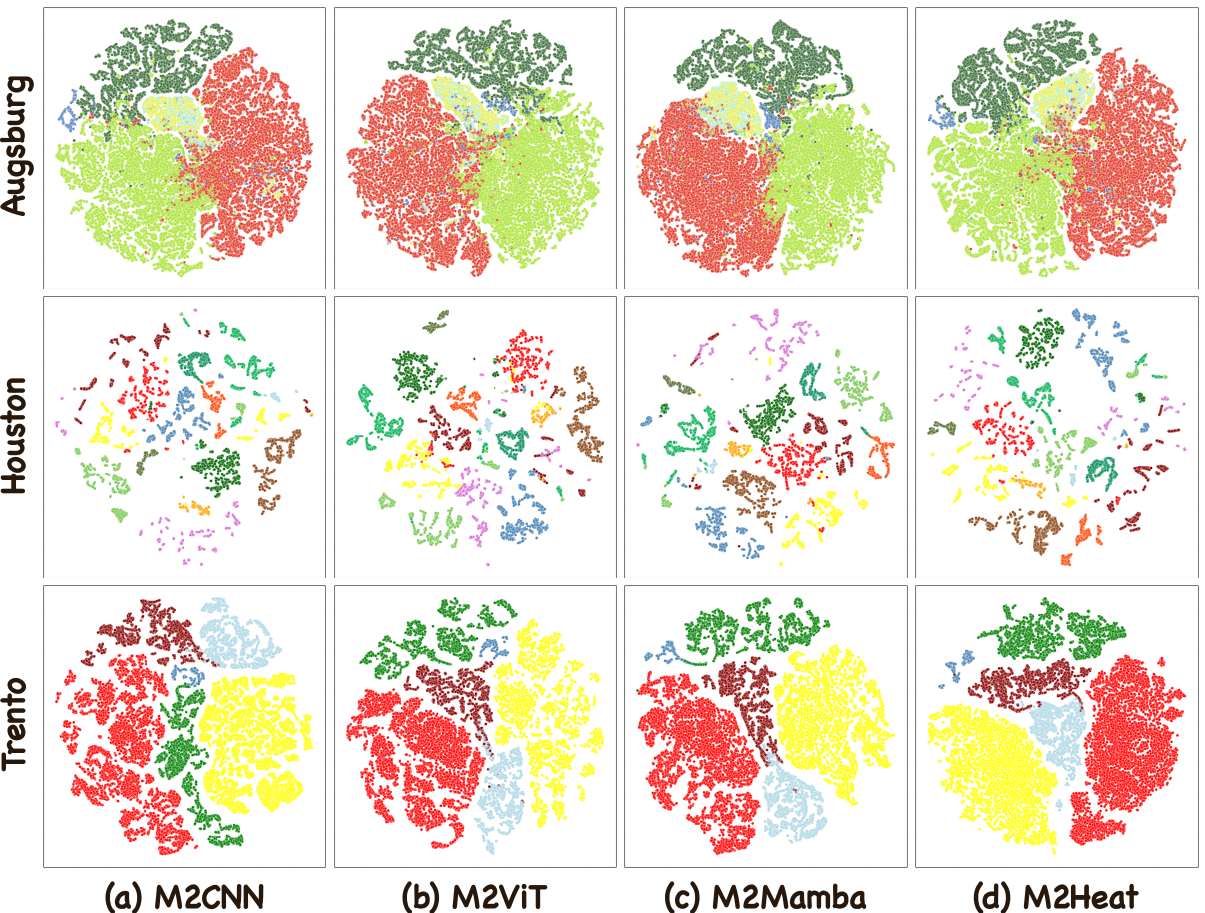}
    \caption{t-SNE visualization on Trento, Houston, and Augsburg datasets. (a) M2CNN: HCO replaced by a convolution. (b) M2ViT: HCO replaced by MSA. (c) M2Mamba: HCO replaced by Mamba. (d) M2Heat: the proposed method.}
    \label{fig:fig9}
\end{figure}

\begin{table}[]
\belowrulesep=0pt
\aboverulesep=0pt
\centering
\caption{OA (\%) under fixed and learnable diffusivity configurations. $S$ denotes a certain number.}
\label{tab:fve_oa}

\footnotesize
\setlength{\tabcolsep}{2.5pt}
\renewcommand{\arraystretch}{1.08}

\begin{tabular}{c|ccccc|ccc}
\toprule
\multirow{2}{*}{\textbf{Dataset}}
& \multicolumn{5}{c|}{\textbf{Fixed FVEs}}
& \multicolumn{3}{c}{\textbf{Learnable FVEs}} \\
\cmidrule(lr){2-6}
\cmidrule(lr){7-9}
& $0.10$ & $0.25$ & $0.50$ & $0.75$ & $1.00$
& $S$ & $P\times P$ & $P\times P\times C$ \\
\midrule
Trento
& 99.44 & 99.31 & 99.27 & 99.24 & 99.21
& 99.27 & 99.42 & \textbf{99.64} \\

Houston
& 88.38 & 87.92 & 88.55 & 88.21 & 88.78
& 87.37 & 88.11 & \textbf{91.20} \\

Augsburg
& 86.92 & 86.11 & 86.56 & 87.01 & 86.73
& 86.64 & 87.09 & \textbf{87.48} \\
\bottomrule
\end{tabular}
\end{table}

\subsection{Ablation Study}
\subsubsection{Contribution of Each Component of M2Heat}
To verify the effectiveness of the key components in M2Heat, we conducted a series of ablation experiments, as summarized in Table \ref{tab:ablation}. S1 is short for the cross-frequency fusion of $U^t_1$ and $U^t_2$, while S2 is short for the dynamic fusion of $U^t_3$ and $\hat{U}^t_f$.

Single-modality baselines using only HSI or LiDAR yield significantly lower accuracies, especially on the Houston and Augsburg datasets, indicating that either modality alone is insufficient for complex urban classification tasks. When features from both modalities are joint learning, performance improves across all datasets, confirming the benefit of multisource fusion.

Introducing S1 to the joint representation consistently enhances OA, with gains of +1.40\%, +0.25\%, and +1.33\% on Trento, Houston, and Augsburg, respectively. This demonstrates the importance of capturing frequency-specific correlations between modalities. Similarly, adding S2 also boosts performance compared to the joint baseline, with notable improvements of +1.34\%, +1.61\%, and +0.67\% on the three datasets, respectively, highlighting the role of dynamic fusion in refining high-level feature integration.

The full M2Heat model, which incorporates both S1 and S2, achieves the highest OA on all datasets: 99.64\%, 91.20\%, and 87.48\%, surpassing all partial configurations. 

Beyond the component-level ablation, we further examine the frequency representations in CFF and spectral operators in HCO. Since the two modules use frequency transforms for different purposes, module-specific alternatives are evaluated separately, as summarized in Table~\ref{tab:transform_ablation}.

For CFF, we compare DCT coefficients, FFT with real--imaginary decomposition (FFT-RI), the proposed amplitude--phase decomposition (FFT-AP), and FFT with low--high frequency decomposition (FFT-L/H). FFT-AP consistently achieves the best performance, outperforming the strongest alternative by 0.13\%, 0.52\%, and 0.42\% on the three datasets. Notably, FFT-RI preserves the same Fourier information but adopts a different representation, while FFT-L/H explicitly separates frequency bands. Their lower accuracies demonstrate the advantage of amplitude--phase decomposition for cross-modal frequency interaction.

For HCO, FFT-, DST-~\cite{yip1980fast}, and DCT-based realizations are compared with GFNet~\cite{rao2021global}, a generic Fourier filtering operator. DCT performs best on Trento and Augsburg and achieves the highest average OA across the three datasets. Although FFT slightly outperforms DCT on Houston2013, the DCT-based realization shows more consistent overall performance and remains aligned with the finite-domain heat-diffusion formulation. The lower performance of DST and GFNet further suggests that the effectiveness of HCO cannot be attributed to generic frequency-domain filtering alone.

Overall, these results validate the distinct frequency designs of M2Heat: FFT-AP is more suitable for cross-modal interaction in CFF, while DCT provides an effective spectral realization of the heat-conduction process in HCO.

\begin{table}[!]
\belowrulesep=0pt
\aboverulesep=0pt
\centering
\caption{Values of the learned effective diffusion coefficients  $k$ under the scalar FVE configuration applied in each dataset.}
\label{tab:fve_1d}

\footnotesize
\setlength{\tabcolsep}{5pt}
\renewcommand{\arraystretch}{1.08}

\begin{tabular}{c||ccc}
\toprule
 & Trento & Houston & Augsburg \\
\midrule \midrule
HSI   & 0.7452 & 1.2356 & 0.7578 \\
LiDAR & 1.2769 & 1.0511 & 0.6903 \\
Fuse  & 18.2251 & 15.1014 & 3.6354 \\
\bottomrule
\end{tabular}
\end{table}

\begin{figure*}
    \centering
    \includegraphics[width=0.9\linewidth]{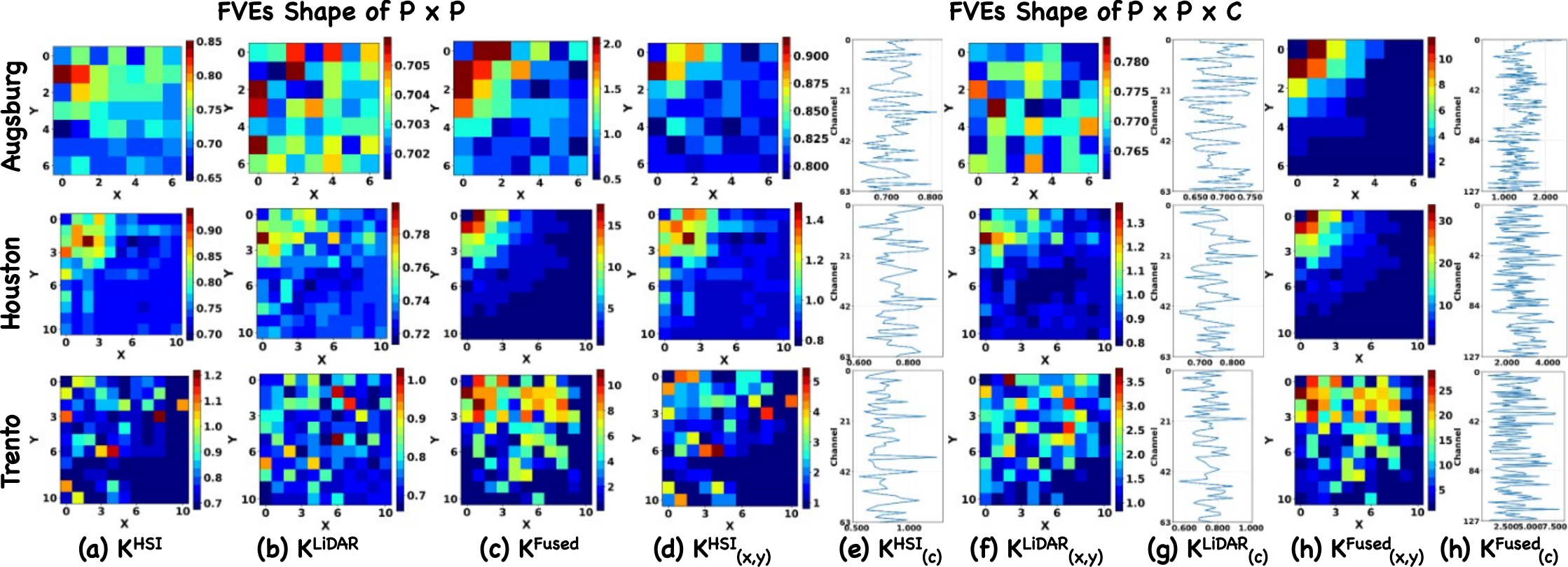}
    \caption{Visualization of different shapes of learnable FVEs on Trento, Houston, and Augsburg datasets.}
    \label{fig:fig10}
\end{figure*}

\begin{figure}
    \centering
    \includegraphics[width=1.0\linewidth]{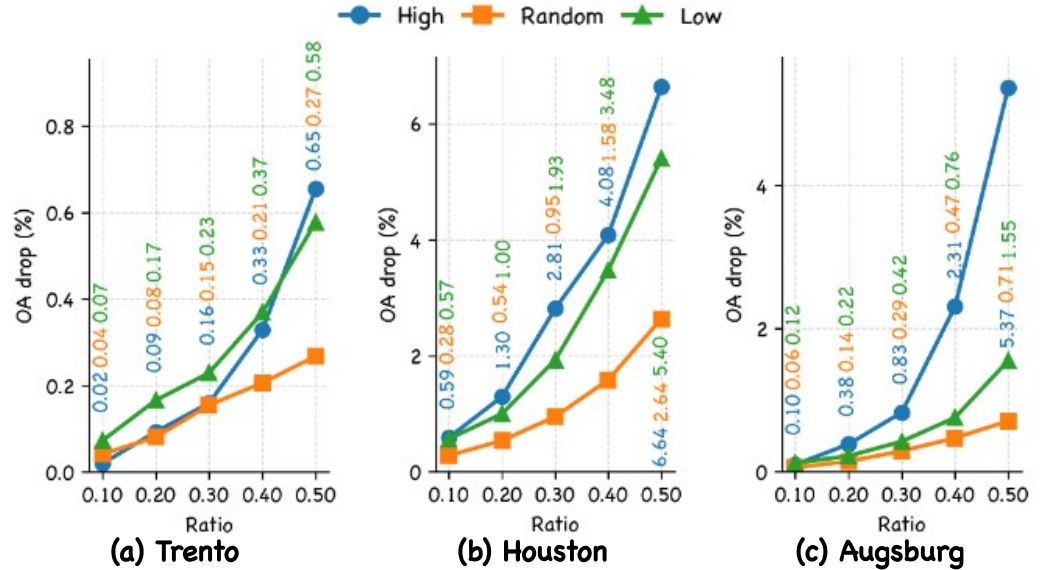}
    \caption{Diffusivity-guided perturbation analysis of HCO on Trento, Houston2013, and Augsburg. High, Random, and Low represent perturbing regions with high diffusivity, random regions, and low diffusivity, respectively.}
    \label{fig:perturbation}
\end{figure}

\subsubsection{Comparison Between Different Operators}

To further validate the effectiveness of the proposed HCO in M2Heat, we compare it with three representative operator designs: $3\times 3$ Conv, Multihead Self Attention (MSA), and Mamba.

As summarized in Table \ref{tab:operator_analysis}, replacing HCO with any of these alternatives leads to noticeable performance degradation. Conv achieves reasonable accuracy on Trento with 99.26\% due to its strong locality modeling, but underperforms on Houston2013 and Augsburg, where large-scale spatial context and inter-modal correlation are crucial. MSA improves global representation capacity, but its computational overhead is prohibitive for large spatial dimensions, and its accuracy on Houston2013 by 87.78\% is notably lower than HCO. Mamba strikes a better balance between complexity and context range; however, it lacks explicit mechanisms for frequency-aware fusion, limiting its discriminative power.

In contrast, HCO attains the highest OA on all datasets: 99.64\%, 91.20\%, and 87.48\%, while maintaining the smallest parameter count by 555.62k and moderate FLOPs with 5.334G. HCO further exhibits the lowest inference time of $0.53$~s and a moderate peak GPU memory footprint of $490.72$~MB, indicating that its frequency-domain computation does not introduce prohibitive runtime or memory overhead.
More importantly, HCO performs two-dimensional global feature propagation in the frequency domain while preserving the spatial organization of image patches, which provides an interpretable alternative to sequence-based global modeling.
This property allows M2Heat to integrate fine-grained local features and long-range dependencies across modalities, leading to favorable classification performance in both homogeneous and heterogeneous scenes.

To assess robustness under limited supervision, we further evaluate the four operators using 25\%, 50\%, and 75\% of the original training samples. For each ratio, the training samples are selected using class-balanced sampling with a fixed random seed to ensure consistent and reproducible subsets. The full-sample result is used as the reference, and the relative OA decay (ROD) is computed as $(\mathrm{OA}_{100}-\mathrm{OA}_{r})/\mathrm{OA}_{100}\times100\%$, where a lower value indicates less performance degradation.

As shown in Table~\ref{tab:sample_robustness}, HCO achieves the highest OA at all reduced training ratios and the lowest average ROD on Trento, Houston, and Augsburg, with 0.252\%, 0.742\%, and 0.490\%, respectively. Although several competing operators obtain lower ROD at individual ratios, HCO shows the smallest overall degradation across all three datasets. This demonstrates that HCO maintains both higher classification accuracy and more stable performance as the available training samples decrease.

To further investigate the discriminative capability of different operators, we conducted a t-Distributed Stochastic Neighbor Embedding (t-SNE) analysis on the output feature representations from the last encoder layer. As shown in Fig.~\ref{fig:fig9}, each point corresponds to a sample, and colors indicate distinct classes, with the color scheme kept consistent with the classification map visualization in Section~\ref{sec:exp}.

On the Trento dataset, the feature clusters produced by M2Heat exhibit the most distinct inter-class boundaries, with minimal overlap, indicating that the proposed HCO effectively enhances class separability in the spectral–spatial feature space. In the more challenging Houston2013 dataset, which contains a larger number of classes with substantial intra-class variability, M2Heat still maintains a clear and compact class distribution in the t-SNE space, reflecting its robustness in complex urban scenes.

For the Augsburg dataset, due to its inherent challenges, such as heterogeneous urban landscapes and subtle spectral differences, most methods, including M2Heat, show partial confusion between Residential Area and Allotment classes. Nevertheless, M2Heat achieves more coherent clustering patterns than the Conv-, MSA-, and Mamba-based baselines, demonstrating its advantage in learning modality-complementary representations with better global discrimination.

\subsubsection{Analysis of Learnable Diffusion Patterns in HCO}

To investigate the effect of the decay coefficient $k$ in the HCO, we compare five fixed, non-learnable constants with three configurations of learnable FVEs: (1) Certain Number, where $k$ is represented as a scalar; (2) $P\times P$, where $k$ varies spatially but remains spectrally invariant; and (3) $P\times P\times C$, where $k$ jointly adapts to spatial positions and spectral channels. The results in Table~\ref{tab:fve_oa} show that the optimal fixed value varies across datasets, whereas the learnable $P\times P\times C$ configuration consistently achieves the highest OA on all three benchmarks. This observation indicates that jointly modeling spatial and channel-dependent diffusion coefficients provides greater flexibility in capturing modality- and scene-specific propagation patterns, thereby motivating its adoption in the final M2Heat architecture.

To further analyze the learned diffusion behavior, Table~\ref{tab:fve_1d} summarizes the scalar FVEs under the Certain Number setting. The HSI and LiDAR branches exhibit moderate and dataset-dependent diffusion strengths, whereas the shared fusion branch consistently learns larger coefficients. This suggests that modality-specific HCO layers mainly preserve heterogeneous cues, while the shared branch requires stronger diffusion for cross-modal interaction and aggregation. The variation across datasets further indicates that HCO adapts its propagation behavior to different scene characteristics.

The $P\times P$ and $P\times P\times C$ FVEs are visualized in Fig.~\ref{fig:fig10}. For the latter, the coefficients are decomposed into channel-averaged spatial maps and channel-wise response curves. Compared with the spatial-only configuration, $P\times P\times C$ exhibits richer spatial and channel variations, while the HSI, LiDAR, and shared branches show distinct response patterns. These observations indicate that HCO adaptively regulates diffusion according to modality-specific and shared representations.

To further quantify whether these patterns are related to model decisions, we conduct a diffusivity-guided perturbation analysis. Spatial positions are ranked by their effective diffusivity and perturbed using three strategies: High selects the largest values, Low selects the smallest values, and Random selects the same number of positions uniformly at random. The HSI, LiDAR, and fused branches are masked independently, with each mask generated according to the distribution of its corresponding effective diffusivity map. The selected positions are replaced by the mean of the unmasked regions in each sample, and Random is averaged over five trials. Perturbation ratios of 10\%, 20\%, 30\%, 40\%, and 50\% are evaluated.

As shown in Fig.~\ref{fig:perturbation}, OA degradation generally increases with the perturbation ratio. Diffusivity-guided perturbations produce larger drops than random perturbation, with high-diffusivity regions becoming particularly sensitive at larger ratios on Houston2013 and Augsburg. Trento shows smaller overall variations but follows a similar tendency at higher perturbation levels. These results establish a quantitative association between the learned diffusivity and prediction sensitivity, complementing the qualitative visualization and supporting the mechanism-level interpretability of HCO.

\section{Conclusion}
In this paper, we introduce M2Heat, a novel multimodal fusion framework inspired by heat conduction principles, specifically designed for HSI--LiDAR joint classification. By integrating modality-specific and shared encoders within a cross-modal conduction-inspired fusion module, M2Heat efficiently captures long-range dependencies with sub-quadratic computational complexity while providing physically interpretable insights into the fusion dynamics. The physics-driven vHeat module, complemented by enhanced FVEs, facilitates precise modeling of anisotropic spectral–spatial heat propagation, thereby enhancing discriminative feature extraction and class separability. A hybrid fusion strategy that synergizes alignment-aware spatial learning with frequency-domain interactions further ensures robust and informative feature integration. Extensive experiments across diverse benchmarks show that M2Heat achieves competitive and favorable overall performance, while the ablation studies verify the contributions of the FVE design and dual-phase fusion mechanism to classification efficacy.

While M2Heat adopts an efficient global modeling strategy through the HCO, its computational cost can still be further reduced compared with linear-complexity operators such as Mamba. This observation motivates our future efforts toward designing more computationally efficient, physics-inspired fusion architectures that retain interpretability while reducing complexity. Furthermore, we aim to enhance robustness and generalization across diverse multimodal RS tasks and extend the applicability of multimodal fusion to cross-scenario and cross-domain settings, paving the way for more practical and scalable RS solutions.

\ifCLASSOPTIONcaptionsoff
  \newpage
\fi

\bibliographystyle{IEEEtran}
\small\bibliography{ref}

\end{document}